\documentclass[10pt,twocolumn,letterpaper]{article}
\usepackage[pagenumbers]{cvpr}
\usepackage{array,tabularx}
\usepackage{listings}
\usepackage{placeins}
\usepackage{microtype}
\definecolor{cvprblue}{rgb}{0.21,0.49,0.74}
\usepackage[pdfusetitle,breaklinks,colorlinks,allcolors=cvprblue]{hyperref}
\graphicspath{{figures/}}
\newcommand{\rev}[1]{#1}
\newcommand{\bluecell}[1]{#1}

\newsavebox{\revisiontablebox}
\lstdefinestyle{PromptBox}{
  basicstyle=\ttfamily\scriptsize,breaklines=true,breakatwhitespace=false,
  columns=fixed,flexiblecolumns=false,frame=single,framerule=0.4pt,
  framesep=4pt,framexleftmargin=0pt,framexrightmargin=0pt,
  xleftmargin=5pt,xrightmargin=5pt,backgroundcolor=\color{gray!8},
  showstringspaces=false,keepspaces=true,aboveskip=0.5em,belowskip=0.5em,
}
\newcommand{\promptrole}[1]{\par\smallskip\noindent\textbf{#1.}\par\nopagebreak[4]\vspace{0.35em}}
\newcolumntype{P}[1]{>{\raggedright\arraybackslash}p{#1}}
\newcommand{\cmark}{\checkmark}
\newcommand{\xmark}{\scalebox{0.85}{$\times$}}
\newcommand{\dataset}{PAPE}
\newcommand{\taxname}{Six-Type Semantic Taxonomy of Object Anomalies}
\newcommand{\tax}{SSTOA}
\newcommand{\real}{\mathrm{R}}
\newcommand{\pl}{\mathrm{P}}
\newcommand{\an}{\mathrm{A}}
\newcommand{\iou}{\operatorname{IoU}}

\newenvironment{resulttable}{\par\captionsetup{type=table}}{\par}
\title{Can You Find the Edit When Nothing Looks Wrong? A Paired Benchmark for Text-Driven Image Manipulation Localization Beyond Semantic Abnormality}
\author{\texorpdfstring{%
Peiyu Zhuang\textsuperscript{1}, Jianquan Yang\textsuperscript{1}, Zhuoying Cai\textsuperscript{1}, Yi Jiang\textsuperscript{1}
Yingquan Chen\textsuperscript{1}, Xinle Wang\textsuperscript{1},\\ Zeqin Yu\textsuperscript{1}, Dongdong Lin\textsuperscript{2}, 
Ruitao Xie\textsuperscript{3}, Kaige Li\textsuperscript{1}, Xiaochun Cao\textsuperscript{1}\\
\textsuperscript{1}Shenzhen Campus of Sun Yat-sen University, China\\
\textsuperscript{2}Huaqiao University, Xiamen, China\\
\textsuperscript{3}Shenzhen University of Advanced Technology, China%
}{Peiyu Zhuang, Jianquan Yang, Zhuoying Cai, Yi Jiang, Yingquan Chen, Xinle Wang, Zeqin Yu, Dongdong Lin, Ruitao Xie, Kaige Li, Xiaochun Cao}}

\begin{document}
\maketitle
\begin{abstract}
\rev{Instruction-driven image editing can alter a photograph while keeping the result coherent with the scene. Existing benchmarks lack matched plausible and anomalous edits of the same target, making it difficult to measure how localization performance changes with semantic plausibility. We construct a paired benchmark by generating semantically plausible and anomalous replacements of the same target in each source photograph. Both edits use the same editor, language, and generation settings. We estimate a separate mask for each output and retain pairs with sufficient mask overlap, enabling direct comparison of localization performance under the two semantic conditions. The release contains 47,747 bilingual triplets from 6,500 photographs, spanning four editors, three source domains, and six intended violation types. Across nine public localizers, editor-averaged IoU is lower on plausible edits in 26 of 27 model--dataset comparisons. This gap persists under pixel-change controls and region-size analyses.These findings reveal a semantic-plausibility shortcut: localization often weakens when an edited result looks ordinary. The benchmark provides a controlled setting for measuring this gap and developing localizers that remain effective on plausible edits.}
\end{abstract}

\section{Introduction}

Instruction-driven image editing can alter what a photograph communicates while keeping the result coherent with the surrounding scene. Recent forensic resources increasingly reflect this shift toward diverse, realistic manipulations and diffusion-based edits~\citep{qu2024modern,wang2025opensdi,zhuang2026sigma}. Such edits create a fundamental challenge for visual verification: a manipulated region may remain physically and semantically plausible, leaving no obvious reason to question its content. A reliable forensic localizer must therefore identify where an image was edited even when the edit does not look wrong.

\begin{figure*}[t]
    \centering
    \includegraphics[width=0.9\textwidth]{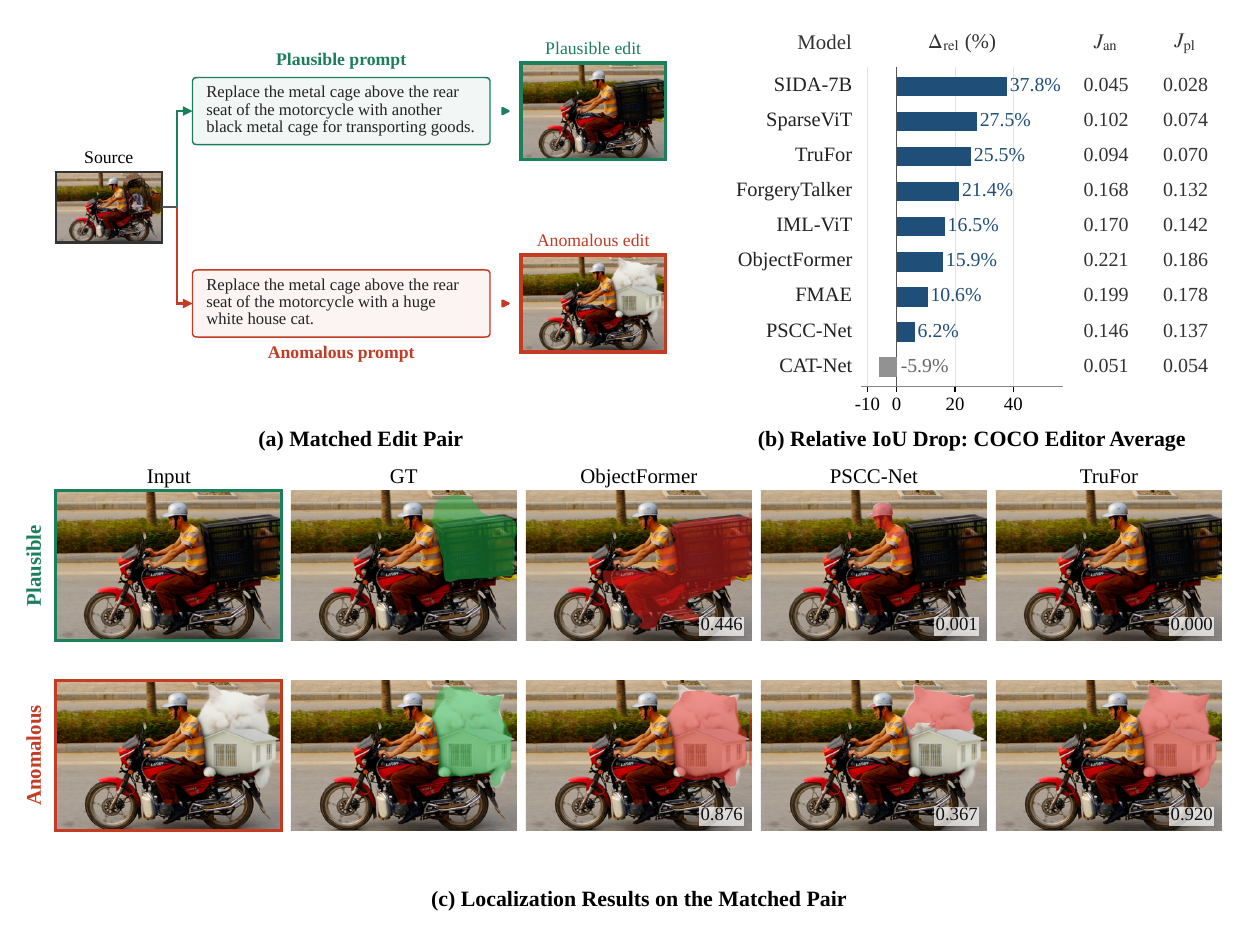}
    \caption{\rev{(a) Matched plausible and anomalous replacements. (b) COCO editor-averaged relative gaps and IoUs for nine localizers, taken directly from Table~\ref{tab:exp-coco}. (c) Predictions on each output, with its independently generated SIGMA\citep{zhuang2026sigma} mask in green and prediction in red. The masks have similar support because the target is shared and pair-mask IoU is at least 0.5.}}
    \label{fig:motivation}
\end{figure*}

\rev{Manipulation localizers are commonly evaluated by pixel-level F1 and IoU averaged over a benchmark. These aggregate scores can conceal whether a model localizes plausible edits as reliably as anomalous ones. We observe substantial accuracy differences between plausible and anomalous replacements of the same target when the source image, editor, and generation settings are held fixed. For the paired edits in Figure~\ref{fig:motivation}(a), ObjectFormer obtains IoU 0.876 on the anomalous replacement and 0.446 on the plausible one (Figure~\ref{fig:motivation}(c)). Both outputs are manipulated, yet the plausible edit is harder to localize. We call this tendency the \emph{semantic-plausibility shortcut}.}

\begin{table}[t]
    \centering
        \caption{\textbf{Construction choices of recent image manipulation localization datasets.} Every listed resource provides pixel masks. Prior datasets evaluate independent images or other diagnostic axes and therefore cannot isolate a same-source plausibility contrast. \dataset{} makes that contrast the evaluation unit. In the last three columns, \cmark{} indicates that the resource satisfies the column criterion and \xmark{} that it does not.}
    \label{tab:positioning}
    \scriptsize
    \setlength{\tabcolsep}{2.2pt}
    \renewcommand{\arraystretch}{1.1}
    \begin{tabularx}{\columnwidth}{@{}lXccc@{}}
    \toprule
    Resource & Forgery process & Masks & \shortstack{Matched\\pair} & \shortstack{Paired\\plausibility} \\
    \midrule
    MIML~\citep{qu2024modern} & Web / manual & \cmark & \xmark & \xmark \\
    GIM~\citep{chen2025gim} & Mask-guided inpaint & \cmark & \xmark & \xmark \\
    OpenSDI~\citep{wang2025opensdi} & Global / local diffusion & \cmark & \xmark & \xmark \\
    LocateEdit~\citep{wu2026locateedit} & Instruction edit & \cmark & \xmark & \xmark \\
    SIGMA~\citep{zhuang2026sigma} & Annotates existing edits & \cmark & \xmark & \xmark \\
    NeXT-IMDL~\citep{li2025nextimdl} & Diverse AIGC edits & \cmark & \xmark & \xmark \\
    HSIM~\citep{zhu2026hsim} & Multi-step sequences & \cmark & \xmark & \xmark \\
    \dataset{} & Instruction edit & \cmark & \cmark & \cmark \\
    \bottomrule
    \end{tabularx}
\end{table}

\rev{Existing benchmarks cover diverse manipulation processes but lack matched plausible and anomalous replacements of the same target (Table~\ref{tab:positioning}). Independent collections mix differences in source content, editors, and edited regions with the plausibility contrast. To examine whether this shortcut recurs across localization methods, we introduce \textbf{Plausible--Anomalous Paired Edits (\dataset{})}. Each triplet combines a source photograph with two edits of the same target. One instruction requests a scene-compatible replacement, and the other requests a specified semantic violation. Both edits use the same editor, instruction language, replacement operation, and generation settings. Each output has its own mask, and overlap filtering retains pairs with comparable edited regions.}

\rev{Our two-stage pipeline generates and validates Chinese and English prompt pairs for six violation types, then applies each pair unchanged across four editors. After SIGMA~\citep{zhuang2026sigma} estimates a separate mask for each output, we filter by mask overlap and align sources across editors. The resulting release contains 47,747 bilingual triplets from 6,500 photographs across COCO, ImageNet, and Places. Across nine public localizers, editor-averaged IoU is lower on plausible edits in 26 of 27 model--dataset comparisons, including eight of nine methods on COCO (Figure~\ref{fig:motivation}(b)).}

\rev{Our contributions are:}
\begin{itemize}
    \item \rev{We identify the semantic-plausibility shortcut and introduce a paired benchmark to measure localization differences between plausible and anomalous replacements.}
    \item \rev{We release the Plausible–Anomalous Paired Edits(\dataset{}) benchmark, with 47,747 bilingual triplets across four editors, three source domains, and six violation types, for developing and evaluating localizers under both semantic conditions.}
    \item \rev{We show that the gap persists in controlled evaluations of nine localization methods using matched edit magnitudes, region-size strata, and alternative masks. }
\end{itemize}

\begin{table*}[!t]
    \centering
    \caption{\textbf{\taxname{} (\tax{}): six sources of incompatibility.} Each row pairs a compatible replacement of the same target with a corresponding semantic violation.}
    \label{tab:taxonomy}
    \scriptsize
    \setlength{\tabcolsep}{4pt}
    \begin{tabular}{@{}llP{0.16\textwidth}P{0.28\textwidth}P{0.28\textwidth}@{}}
    \toprule
    ID & Locus & Category & Defining question & Contrast \\
    \midrule
    C1 & Object & Physical attribute & Does material or physical state violate ordinary object properties? & Rigid ceramic mug / mug made of unsupported liquid \\
    C2 & Object & Function or affordance & Does the object support its depicted use? & Functional watering can / watering can with a sealed spout being used to pour \\
    C3 & Object & Biological structure & Are anatomy, part count, or bodily configuration inconsistent? & Cat with ordinary anatomy / cat with an extra head \\
    C4 & Relation & Spatial relation & Are support, contact, containment, or scale inconsistent? & Book resting on a shelf / book intersecting the solid shelf \\
    C5 & Relation & Interaction relation & Is an action or relation between entities incoherent? & Hand grasping a cup handle / hand passing through the solid handle \\
    C6 & Scene & Scene context & Is an otherwise recognizable object incompatible with its surroundings? & Chair at a dining table / traffic barrier replacing the chair at a formal table \\
    \bottomrule
    \end{tabular}
\end{table*}

\section{Related Work}

\paragraph{Datasets for image manipulation localization.}
\rev{MIML~\citep{qu2024modern} collects manual forgeries, while GIM~\citep{chen2025gim}, OpenSDI~\citep{wang2025opensdi}, and DEAL-300K~\citep{zhang2025deal} extend mask supervision to generative edits. SIGMA~\citep{zhuang2026sigma} annotates existing instruction-editing records. LocateEditBench~\citep{wu2026locateedit} and PromptForge-350k~\citep{wang2026promptforge} construct instruction-editing benchmarks. NeXT-IMDL~\citep{li2025nextimdl} tests editor, semantic, and granularity generalization, and HSIM~\citep{zhu2026hsim} studies hierarchical editing trajectories. These resources score individual images or other diagnostic axes. \dataset{} pairs plausible and anomalous replacements of the same source and target under one editor (Table~\ref{tab:positioning}).}

\paragraph{Forensic and semantic representations.}
\rev{Existing localizers draw on forensic cues, visual representations, or multimodal reasoning. Forensic-cue methods exploit compression (CAT-Net~\citep{kwon2021catnet}), spatio-channel correlations (PSCC-Net~\citep{liu2021pscc}), and RGB--noise fusion (TruFor~\citep{guillaro2023trufor}). Visual-representation methods include ObjectFormer~\citep{wang2022objectformer} for object--patch consistency and IML-ViT~\citep{ma2023imlvit} for high-resolution features, trained with mask or boundary supervision for localization accuracy. SparseViT~\citep{su2025sparsevit} uses sparse attention, while FMAE~\citep{zhu2025fmae} combines masked-autoencoder features and noise extraction. Multimodal methods SIDA-7B~\citep{huang2025sida} and ForgeryTalker~\citep{lian2026forgerytalker} add explanation supervision for interpretability. In this paper, we propose \dataset{} to examine how localization varies across edit semantics through matched plausible--anomalous comparisons, exposing gaps hidden by aggregate accuracy and enabling condition-specific evaluation of future improvements.}

\section{Defining the Paired Plausibility Test}
\label{sec:definition}

The paired plausibility test asks whether localization quality changes when the same target is edited under two intended semantic conditions. One replacement is compatible with the scene, whereas the other is designed to violate a specified physical, biological, functional, spatial, interaction, or contextual expectation. The comparison is between a plausible edit and an anomalous edit, and the semantic condition is recorded separately from the manipulation label.

\paragraph{Structuring the Semantic Contrast.}
\rev{The \taxname{} (\tax{}) specifies six kinds of intended conflict (Table~\ref{tab:taxonomy}). Categories C1--C3 concern object properties, C4--C5 concern relations, and C6 concerns scene context. For a fixed source and target, $P_i^\pl$ requests a compatible replacement and $P_i^\an$ requests the sampled violation $c_i$. Categories describe the kind of conflict rather than an ordered abnormality level. }

\paragraph{Paired examples.}
Each triplet contains a source photograph $I_i^\real$ and two replacements of the same target, $I_i^\pl$ and $I_i^\an$, produced by editor $e_i$ under shared settings $\theta_i$. For $z\in\{\pl,\an\}$, each output has its own mask $M_i^z$. Both edits have manipulation label $y=1$, with intended anomaly labels $a^\pl=0$ and $a^\an=1$. The source has $y=0$. Appendix~\ref{app:pair_record} specifies the full record.

\section{\rev{Design and Construction of the proposed benchmark}}
\label{sec:construction}

\subsection{Stage A: Prompt Pair Generation}

\rev{Stage~A samples a violation type and uses Qwen3.5-9B~\citep{qwen35blog} for perception and paired instruction generation, with separate structured templates (details are shown in Appendix~\ref{app:prompt_perception} and Appendix~\ref{app:prompt_pair}). The perception module identifies visible targets and regions. The generation module formulates Chinese and English plausible--anomalous instructions as replacements of the same target, thereby keeping the edit operation consistent within each pair, and records semantic rationales and pair-control notes. Structured Prompt Pair Validation (SPPV) then verifies JSON parseability, the presence of required bilingual fields, and the use of replacement-form instructions before editing (Appendix~\ref{app:prompt_validation}).} 

\begin{figure*}[t]
    \centering
    \includegraphics[width=0.92\textwidth,height=0.38\textheight,keepaspectratio]{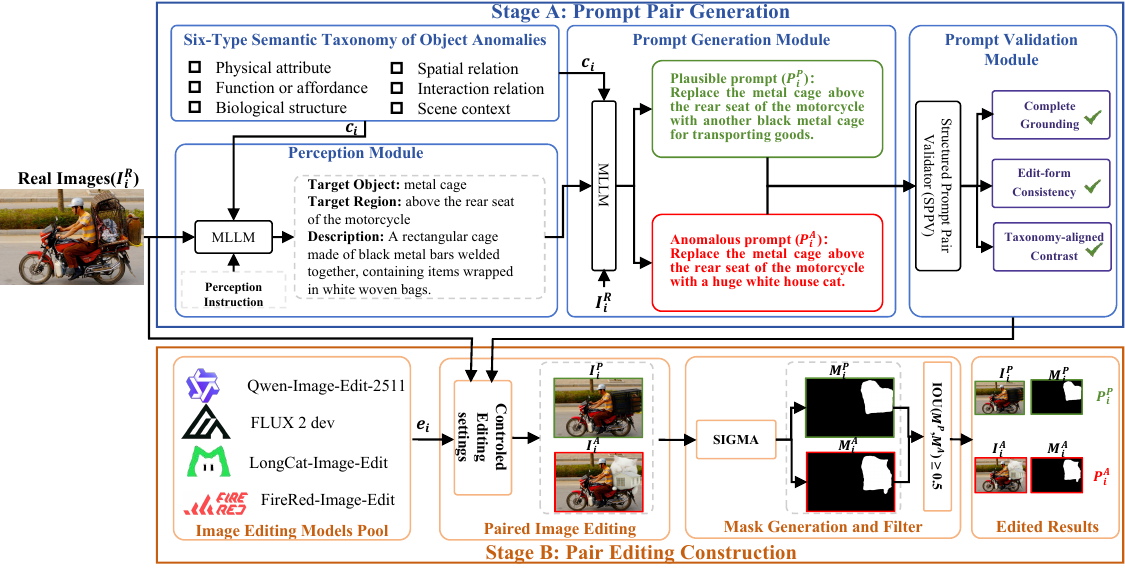}
    \caption{\rev{Two-stage pair construction. Stage~A generates bilingual replacements of a shared target and checks instruction structure. Stage~B applies each language's unchanged prompt pair to all four editors, with matched settings within each editor. SIGMA estimates a separate mask for each output, and pair-mask IoU must be at least 0.5.}}
    \label{fig:pipeline}
\end{figure*}

\begin{figure*}[!t]
    \centering
    \includegraphics[width=\textwidth]{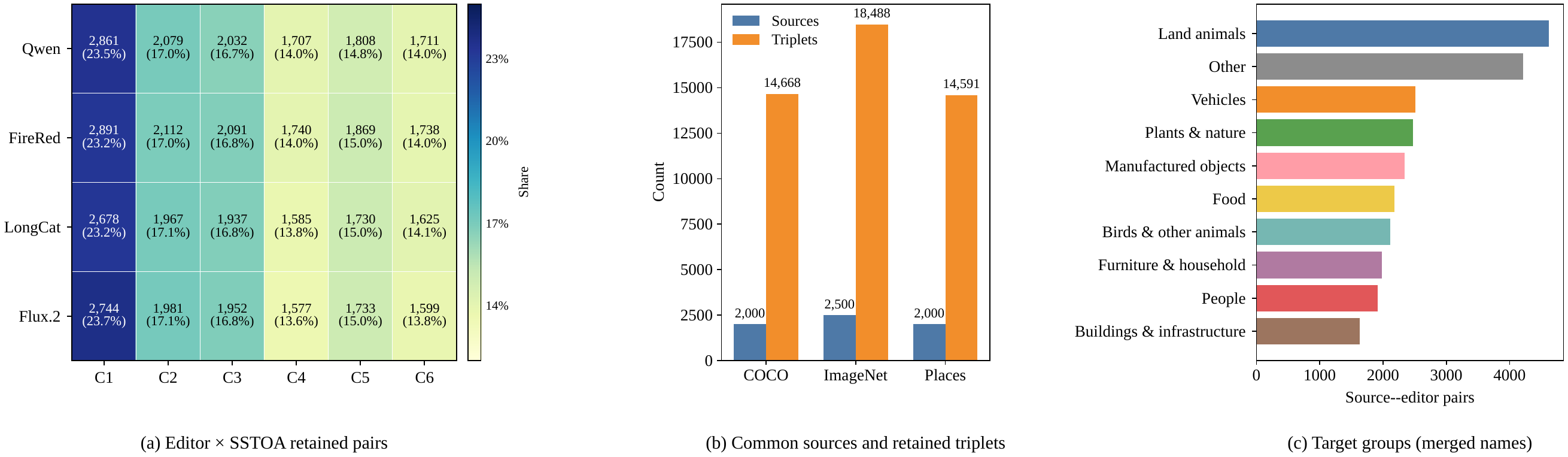}
    \caption{Distribution of the final \dataset{} release. (a) Triplet counts and within-editor shares for the six intended violation types. (b) Source photographs and retained triplets by dataset. (c) Target-object groups, counted over 26,000 source--editor combinations.}
    \label{fig:coverage}
\end{figure*}

\subsection{\rev{Stage B: Paired Editing and Mask Generation}}

\rev{For each source and language, \textbf{the same instruction pair is applied unchanged to all four editors}: Qwen-Image-Edit~\citep{wu2025qwenimage}, FireRed-Image-Edit~\citep{qiao2026firered}, LongCat-Image-Edit~\citep{longcat2025image}, and Flux.2-dev~\citep{blackforest2025flux2}. These public checkpoints support instruction-guided replacement and vary in scale and conditioning architecture (Appendix~\ref{app:editor_design}). Within each editor, plausible and anomalous generations share the editing configuration $\theta_i$. Chinese and English broaden instruction coverage and provide traceable language variants of the same intended contrast. Each language produces separate images and masks, and comparisons are within language.} \emph{Mask generation and filtering} invokes SIGMA~\citep{zhuang2026sigma} on each edited image with its instruction to estimate $M_i^{\pl}$ and $M_i^{\an}$. Empty or failed masks are discarded. A record is kept only when the two mask supports align:
\begin{equation}
q_i=\iou(M_i^\pl,M_i^\an)
=\frac{|M_i^\pl\cap M_i^\an|}{|M_i^\pl\cup M_i^\an|}\geq 0.5.
\label{eq:filter}
\end{equation}
The resulting release unit contains $(I_i^{\pl},M_i^{\pl})$ and $(I_i^{\an},M_i^{\an})$ for the same source, editor, and $\theta_i$, so downstream localization compares plausible and anomalous conditions on comparable regions.

\begin{figure*}[!t]\centering
    \includegraphics[width=\textwidth]{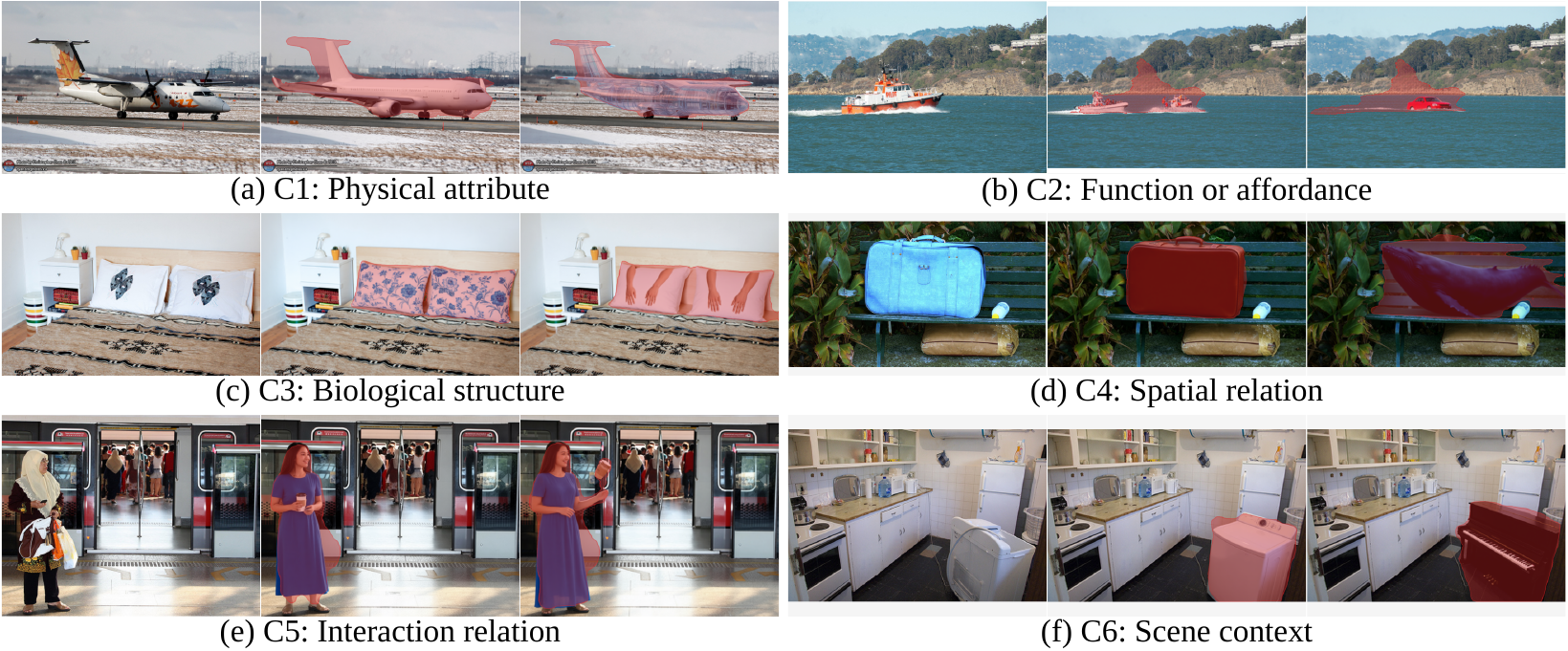}
    \caption{\rev{Six intended violation types in retained English COCO pairs. Each labeled group shows the source, plausible output, and anomalous output from left to right, with the corresponding output-specific mask overlays. Category IDs follow Table~\ref{tab:taxonomy}.}}\label{fig:gallery}
\end{figure*}

\subsection{Dataset Distribution}
\label{sec:dataset_distribution}
The pipeline produces 47,747 language-specific triplets from 6,500 photographs, containing 95,494 edited images with corresponding masks. Figure~\ref{fig:coverage} shows the distribution of intended violation types, source domains, and target objects.

The six violation types have similar proportions across editors (Figure~\ref{fig:coverage}(a)). Physical-attribute violations (C1) account for 23.2\%--23.7\% of each editor's triplets, and spatial-relation violations (C4) account for 13.6\%--14.0\%. Figure~\ref{fig:coverage}(b) shows 14,668 triplets from 2,000 COCO photographs, 18,488 from 2,500 ImageNet photographs, and 14,591 from 2,000 Places photographs. Figure~\ref{fig:coverage}(c) merges 3,483 target names into semantic groups. Land animals contribute 4,624 source--editor combinations. The other named groups cover vehicles, plants, manufactured objects, food, birds, furniture, people, and buildings. Unmatched names fall under Other. To further visualize the dataset's controlled editing outcomes, Figure~\ref{fig:gallery} presents representative source--plausible--anomalous triplets for all six violation types, together with their output-specific mask overlays.

\raggedbottom
\section{Experiments}
\label{sec:experiments}
\renewcommand{\dbltopfraction}{0.75}
\setlength{\textfloatsep}{10pt plus 2pt minus 2pt}
\setlength{\dbltextfloatsep}{10pt plus 2pt minus 2pt}
\setlength{\floatsep}{8pt plus 2pt minus 2pt}
\setlength{\dblfloatsep}{8pt plus 2pt minus 2pt}

\begin{table*}[!t]
\centering
\caption{\rev{Paired localization on the official COCO test split.}}
\label{tab:cpse-zs-coco}
\label{tab:exp-coco}
\scriptsize
\setlength{\tabcolsep}{2.2pt}
\resizebox{0.9\textwidth}{!}{%
\begin{tabular}{lccccccccccccccc}
\toprule
    & \multicolumn{3}{c}{Qwen} & \multicolumn{3}{c}{FireRed} & \multicolumn{3}{c}{LongCat} & \multicolumn{3}{c}{Flux.2} & \multicolumn{3}{c}{Avg} \\
\cmidrule(lr){2-4}\cmidrule(lr){5-7}\cmidrule(lr){8-10}\cmidrule(lr){11-13}\cmidrule(lr){14-16}
Method & $\bar{J}_{\pl}$ & $\bar{J}_{\an}$ & $\Delta_{\mathrm{rel}}$ & $\bar{J}_{\pl}$ & $\bar{J}_{\an}$ & $\Delta_{\mathrm{rel}}$ & $\bar{J}_{\pl}$ & $\bar{J}_{\an}$ & $\Delta_{\mathrm{rel}}$ & $\bar{J}_{\pl}$ & $\bar{J}_{\an}$ & $\Delta_{\mathrm{rel}}$ & $\bar{J}_{\pl}$ & $\bar{J}_{\an}$ & $\Delta_{\mathrm{rel}}$ \\
\midrule
CAT-Net~\bluecell{(WACV2021)} & 0.072 & 0.060 & -0.200 & 0.046 & 0.053 & 0.132 & 0.032 & 0.032 & 0.000 & 0.065 & 0.058 & -0.121 & 0.054 & 0.051 & -0.059 \\
ObjectFormer~\bluecell{(CVPR2022)} & 0.207 & 0.255 & 0.188 & 0.190 & 0.223 & 0.148 & 0.185 & 0.210 & 0.119 & 0.160 & 0.196 & 0.184 & 0.186 & 0.221 & 0.159 \\
PSCC-Net~\bluecell{(TCSVT2022)} & 0.144 & 0.150 & 0.040 & 0.157 & 0.161 & 0.025 & 0.092 & 0.119 & 0.227 & 0.156 & 0.153 & -0.020 & 0.137 & 0.146 & 0.062 \\
IML-ViT~\bluecell{(arXiv2023)} & 0.155 & 0.215 & 0.279 & 0.163 & 0.200 & 0.185 & 0.094 & 0.121 & 0.223 & 0.155 & 0.143 & -0.084 & 0.142 & 0.170 & 0.165 \\
TruFor~\bluecell{(CVPR2023)} & 0.101 & 0.144 & 0.299 & 0.101 & 0.143 & 0.294 & 0.027 & 0.034 & 0.206 & 0.052 & 0.057 & 0.088 & 0.070 & 0.094 & 0.255 \\
SIDA-7B~\bluecell{(CVPR2025)} & 0.036 & 0.066 & 0.455 & 0.026 & 0.045 & 0.422 & 0.034 & 0.054 & 0.370 & 0.016 & 0.016 & 0.000 & 0.028 & 0.045 & 0.378 \\
\bluecell{SparseViT (AAAI2025)} & \bluecell{0.098} & \bluecell{0.154} & \bluecell{0.364} & \bluecell{0.094} & \bluecell{0.128} & \bluecell{0.266} & \bluecell{0.025} & \bluecell{0.037} & \bluecell{0.324} & \bluecell{0.078} & \bluecell{0.089} & \bluecell{0.124} & \bluecell{0.074} & \bluecell{0.102} & \bluecell{0.275} \\
\bluecell{FMAE (AAAI2025)} & \bluecell{0.205} & \bluecell{0.238} & \bluecell{0.139} & \bluecell{0.194} & \bluecell{0.210} & \bluecell{0.076} & \bluecell{0.125} & \bluecell{0.150} & \bluecell{0.167} & \bluecell{0.187} & \bluecell{0.196} & \bluecell{0.046} & \bluecell{0.178} & \bluecell{0.199} & \bluecell{0.106} \\
ForgeryTalker~\bluecell{(ACL2026)} & 0.119 & 0.167 & 0.287 & 0.126 & 0.164 & 0.232 & 0.199 & 0.224 & 0.112 & 0.084 & 0.118 & 0.288 & 0.132 & 0.168 & 0.214 \\
\bottomrule
\end{tabular}}
\end{table*}

\begin{table*}[!t]
\centering
\caption{\rev{Paired localization on the official ImageNet test split.}}
\label{tab:cpse-zs-imagenet}
\label{tab:exp-imagenet}
\scriptsize
\setlength{\tabcolsep}{2.2pt}
\resizebox{0.9\textwidth}{!}{%
\begin{tabular}{lccccccccccccccc}
\toprule
    & \multicolumn{3}{c}{Qwen} & \multicolumn{3}{c}{FireRed} & \multicolumn{3}{c}{LongCat} & \multicolumn{3}{c}{Flux.2} & \multicolumn{3}{c}{Avg} \\
\cmidrule(lr){2-4}\cmidrule(lr){5-7}\cmidrule(lr){8-10}\cmidrule(lr){11-13}\cmidrule(lr){14-16}
Method & $\bar{J}_{\pl}$ & $\bar{J}_{\an}$ & $\Delta_{\mathrm{rel}}$ & $\bar{J}_{\pl}$ & $\bar{J}_{\an}$ & $\Delta_{\mathrm{rel}}$ & $\bar{J}_{\pl}$ & $\bar{J}_{\an}$ & $\Delta_{\mathrm{rel}}$ & $\bar{J}_{\pl}$ & $\bar{J}_{\an}$ & $\Delta_{\mathrm{rel}}$ & $\bar{J}_{\pl}$ & $\bar{J}_{\an}$ & $\Delta_{\mathrm{rel}}$ \\
\midrule
CAT-Net~\bluecell{(WACV2021)} & 0.072 & 0.079 & 0.089 & 0.065 & 0.069 & 0.058 & 0.045 & 0.048 & 0.063 & 0.056 & 0.066 & 0.152 & 0.059 & 0.066 & 0.106 \\
ObjectFormer~\bluecell{(CVPR2022)} & 0.203 & 0.228 & 0.110 & 0.170 & 0.192 & 0.115 & 0.175 & 0.199 & 0.121 & 0.146 & 0.161 & 0.093 & 0.174 & 0.195 & 0.108 \\
PSCC-Net~\bluecell{(TCSVT2022)} & 0.163 & 0.175 & 0.069 & 0.193 & 0.181 & -0.066 & 0.120 & 0.131 & 0.084 & 0.159 & 0.166 & 0.042 & 0.159 & 0.163 & 0.025 \\
IML-ViT~\bluecell{(arXiv2023)} & 0.174 & 0.228 & 0.237 & 0.194 & 0.208 & 0.067 & 0.111 & 0.131 & 0.153 & 0.145 & 0.138 & -0.051 & 0.156 & 0.176 & 0.114 \\
TruFor~\bluecell{(CVPR2023)} & 0.135 & 0.166 & 0.187 & 0.114 & 0.128 & 0.109 & 0.033 & 0.045 & 0.267 & 0.037 & 0.046 & 0.196 & 0.080 & 0.096 & 0.167 \\
SIDA-7B~\bluecell{(CVPR2025)} & 0.022 & 0.042 & 0.476 & 0.024 & 0.034 & 0.294 & 0.033 & 0.045 & 0.267 & 0.009 & 0.013 & 0.308 & 0.022 & 0.034 & 0.353 \\
\bluecell{SparseViT (AAAI2025)} & \bluecell{0.138} & \bluecell{0.168} & \bluecell{0.179} & \bluecell{0.125} & \bluecell{0.126} & \bluecell{0.008} & \bluecell{0.028} & \bluecell{0.037} & \bluecell{0.243} & \bluecell{0.068} & \bluecell{0.079} & \bluecell{0.139} & \bluecell{0.090} & \bluecell{0.103} & \bluecell{0.126} \\
\bluecell{FMAE (AAAI2025)} & \bluecell{0.323} & \bluecell{0.363} & \bluecell{0.110} & \bluecell{0.304} & \bluecell{0.315} & \bluecell{0.035} & \bluecell{0.217} & \bluecell{0.228} & \bluecell{0.048} & \bluecell{0.247} & \bluecell{0.247} & \bluecell{0.000} & \bluecell{0.273} & \bluecell{0.288} & \bluecell{0.052} \\
ForgeryTalker~\bluecell{(ACL2026)} & 0.195 & 0.252 & 0.226 & 0.190 & 0.241 & 0.212 & 0.272 & 0.301 & 0.096 & 0.120 & 0.159 & 0.245 & 0.194 & 0.238 & 0.185 \\
\bottomrule
\end{tabular}}
\end{table*}

\begin{table*}[!t]
\centering
\caption{\rev{Paired localization on the official Places test split.}}
\label{tab:cpse-zs-places}
\label{tab:exp-places}
\scriptsize
\setlength{\tabcolsep}{2.2pt}
\resizebox{0.9\textwidth}{!}{%
\begin{tabular}{lccccccccccccccc}
\toprule
    & \multicolumn{3}{c}{Qwen} & \multicolumn{3}{c}{FireRed} & \multicolumn{3}{c}{LongCat} & \multicolumn{3}{c}{Flux.2} & \multicolumn{3}{c}{Avg} \\
\cmidrule(lr){2-4}\cmidrule(lr){5-7}\cmidrule(lr){8-10}\cmidrule(lr){11-13}\cmidrule(lr){14-16}
Method & $\bar{J}_{\pl}$ & $\bar{J}_{\an}$ & $\Delta_{\mathrm{rel}}$ & $\bar{J}_{\pl}$ & $\bar{J}_{\an}$ & $\Delta_{\mathrm{rel}}$ & $\bar{J}_{\pl}$ & $\bar{J}_{\an}$ & $\Delta_{\mathrm{rel}}$ & $\bar{J}_{\pl}$ & $\bar{J}_{\an}$ & $\Delta_{\mathrm{rel}}$ & $\bar{J}_{\pl}$ & $\bar{J}_{\an}$ & $\Delta_{\mathrm{rel}}$ \\
\midrule
CAT-Net~\bluecell{(WACV2021)} & 0.072 & 0.127 & 0.433 & 0.063 & 0.086 & 0.267 & 0.030 & 0.047 & 0.362 & 0.057 & 0.077 & 0.260 & 0.056 & 0.084 & 0.333 \\
ObjectFormer~\bluecell{(CVPR2022)} & 0.161 & 0.231 & 0.303 & 0.132 & 0.185 & 0.286 & 0.151 & 0.192 & 0.214 & 0.134 & 0.191 & 0.298 & 0.145 & 0.200 & 0.275 \\
PSCC-Net~\bluecell{(TCSVT2022)} & 0.117 & 0.196 & 0.403 & 0.130 & 0.164 & 0.207 & 0.072 & 0.100 & 0.280 & 0.125 & 0.140 & 0.107 & 0.111 & 0.150 & 0.260 \\
IML-ViT~\bluecell{(arXiv2023)} & 0.169 & 0.273 & 0.381 & 0.155 & 0.187 & 0.171 & 0.079 & 0.116 & 0.319 & 0.117 & 0.139 & 0.158 & 0.130 & 0.179 & 0.274 \\
TruFor~\bluecell{(CVPR2023)} & 0.217 & 0.322 & 0.326 & 0.140 & 0.183 & 0.235 & 0.027 & 0.053 & 0.491 & 0.037 & 0.067 & 0.448 & 0.105 & 0.156 & 0.327 \\
SIDA-7B~\bluecell{(CVPR2025)} & 0.033 & 0.067 & 0.507 & 0.032 & 0.051 & 0.373 & 0.027 & 0.043 & 0.372 & 0.012 & 0.019 & 0.368 & 0.026 & 0.045 & 0.422 \\
\bluecell{SparseViT (AAAI2025)} & \bluecell{0.196} & \bluecell{0.301} & \bluecell{0.349} & \bluecell{0.148} & \bluecell{0.167} & \bluecell{0.114} & \bluecell{0.023} & \bluecell{0.037} & \bluecell{0.378} & \bluecell{0.061} & \bluecell{0.075} & \bluecell{0.187} & \bluecell{0.107} & \bluecell{0.145} & \bluecell{0.262} \\
\bluecell{FMAE (AAAI2025)} & \bluecell{0.237} & \bluecell{0.312} & \bluecell{0.240} & \bluecell{0.205} & \bluecell{0.245} & \bluecell{0.163} & \bluecell{0.136} & \bluecell{0.161} & \bluecell{0.155} & \bluecell{0.180} & \bluecell{0.212} & \bluecell{0.151} & \bluecell{0.190} & \bluecell{0.232} & \bluecell{0.181} \\
ForgeryTalker~\bluecell{(ACL2026)} & 0.120 & 0.170 & 0.294 & 0.140 & 0.182 & 0.231 & 0.182 & 0.221 & 0.176 & 0.095 & 0.131 & 0.275 & 0.134 & 0.176 & 0.239 \\
\bottomrule
\end{tabular}}
\end{table*}

\begin{table*}[t]
\centering
\begin{minipage}[t]{\dimexpr(\textwidth-\columnsep)/2\relax}
\vspace{0pt}
\begin{resulttable}
\centering
\caption{Pixel-change balance on official-test pairs. P/A denote plausible/anomalous means. MAD is the mean absolute paired difference.}
\label{tab:control-balance}
\begin{tabular}{@{}lccc@{}}\toprule
Subset & Retained & Mean P / A & MAD \\\midrule
Base & 100.0\% & 0.2636 / 0.2688 & 0.0782 \\
N & 99.5\% & 0.2648 / 0.2699 & 0.0784 \\
S8 & 51.3\% & 0.2602 / 0.2614 & 0.0261 \\
S9 & 28.4\% & 0.2585 / 0.2591 & 0.0127 \\
\bottomrule\end{tabular}
\end{resulttable}
\vspace{\floatsep}

\begin{resulttable}
\centering\footnotesize
\caption{Pixel-change matched localization ($s_{\rm pix}\geq0.9$, 28.4\% retained).}
\label{tab:matched-localization}
\setlength{\tabcolsep}{1pt}
\begin{tabular*}{\columnwidth}{@{\extracolsep{\fill}}l*{9}{r}@{}}\toprule
Method & \multicolumn{3}{c}{COCO} & \multicolumn{3}{c}{ImageNet} & \multicolumn{3}{c}{Places} \\
\cmidrule(lr){2-4}\cmidrule(lr){5-7}\cmidrule(lr){8-10}
 & $J_\pl$ & $J_\an$ & $\Delta_{\rm rel}$ & $J_\pl$ & $J_\an$ & $\Delta_{\rm rel}$ & $J_\pl$ & $J_\an$ & $\Delta_{\rm rel}$ \\\midrule
CAT-Net & 0.05 & 0.05 & $-0.03$ & 0.06 & 0.07 & $+0.14$ & 0.05 & 0.08 & $+0.34$ \\
ObjectFormer & 0.17 & 0.20 & $+0.16$ & 0.16 & 0.18 & $+0.10$ & 0.14 & 0.19 & $+0.25$ \\
PSCC-Net & 0.15 & 0.13 & $-0.11$ & 0.16 & 0.17 & $+0.02$ & 0.11 & 0.14 & $+0.21$ \\
IML-ViT & 0.14 & 0.17 & $+0.20$ & 0.15 & 0.17 & $+0.11$ & 0.13 & 0.17 & $+0.27$ \\
TruFor & 0.08 & 0.10 & $+0.16$ & 0.08 & 0.10 & $+0.17$ & 0.10 & 0.15 & $+0.31$ \\
SIDA-7B & 0.02 & 0.04 & $+0.38$ & 0.02 & 0.03 & $+0.33$ & 0.02 & 0.04 & $+0.49$ \\
SparseViT & 0.09 & 0.11 & $+0.20$ & 0.09 & 0.11 & $+0.14$ & 0.10 & 0.14 & $+0.29$ \\
FMAE & 0.18 & 0.20 & $+0.11$ & 0.26 & 0.28 & $+0.06$ & 0.18 & 0.23 & $+0.21$ \\
ForgeryTalker & 0.14 & 0.17 & $+0.18$ & 0.19 & 0.23 & $+0.18$ & 0.14 & 0.17 & $+0.22$ \\
\bottomrule\end{tabular*}
\end{resulttable}
\end{minipage}\hfill%
\begin{minipage}[t]{\dimexpr(\textwidth-\columnsep)/2\relax}
\vspace{0pt}
\begin{resulttable}
\centering\footnotesize
\caption{Mask-area matched localization ($s_{\rm area}\geq0.9$, 42.0\% retained).}
\label{tab:area-matched-localization}
\setlength{\tabcolsep}{1pt}
\begin{tabular*}{\columnwidth}{@{\extracolsep{\fill}}l*{9}{r}@{}}\toprule
Method & \multicolumn{3}{c}{COCO} & \multicolumn{3}{c}{ImageNet} & \multicolumn{3}{c}{Places} \\
\cmidrule(lr){2-4}\cmidrule(lr){5-7}\cmidrule(lr){8-10}
 & $J_\pl$ & $J_\an$ & $\Delta_{\rm rel}$ & $J_\pl$ & $J_\an$ & $\Delta_{\rm rel}$ & $J_\pl$ & $J_\an$ & $\Delta_{\rm rel}$ \\\midrule
CAT-Net & 0.06 & 0.05 & $-0.31$ & 0.06 & 0.06 & $+0.12$ & 0.06 & 0.08 & $+0.26$ \\
ObjectFormer & 0.19 & 0.21 & $+0.10$ & 0.17 & 0.19 & $+0.08$ & 0.14 & 0.18 & $+0.22$ \\
PSCC-Net & 0.14 & 0.14 & $+0.03$ & 0.16 & 0.16 & $+0.02$ & 0.11 & 0.14 & $+0.19$ \\
IML-ViT & 0.13 & 0.16 & $+0.22$ & 0.15 & 0.17 & $+0.12$ & 0.12 & 0.16 & $+0.22$ \\
TruFor & 0.07 & 0.10 & $+0.29$ & 0.08 & 0.09 & $+0.14$ & 0.10 & 0.14 & $+0.32$ \\
SIDA-7B & 0.03 & 0.05 & $+0.47$ & 0.02 & 0.03 & $+0.29$ & 0.03 & 0.04 & $+0.42$ \\
SparseViT & 0.07 & 0.11 & $+0.33$ & 0.09 & 0.10 & $+0.10$ & 0.10 & 0.13 & $+0.25$ \\
FMAE & 0.17 & 0.19 & $+0.11$ & 0.27 & 0.29 & $+0.07$ & 0.18 & 0.22 & $+0.18$ \\
ForgeryTalker & 0.15 & 0.18 & $+0.19$ & 0.21 & 0.25 & $+0.16$ & 0.15 & 0.18 & $+0.19$ \\
\bottomrule\end{tabular*}
\end{resulttable}
\vspace{\floatsep}
\begin{resulttable}
\centering\footnotesize
\caption{Localization on realized (372 pairs) and unrealized (128 pairs) violations.}
\label{tab:realization}
\setlength{\tabcolsep}{1pt}
\begin{tabular*}{\columnwidth}{@{\extracolsep{\fill}}l*{6}{r}@{}}\toprule
Method & \multicolumn{3}{c}{Realized} & \multicolumn{3}{c}{Unrealized} \\
\cmidrule(lr){2-4}\cmidrule(lr){5-7}
 & $J_\pl$ & $J_\an$ & $\Delta_{\rm rel}$ & $J_\pl$ & $J_\an$ & $\Delta_{\rm rel}$ \\\midrule
CAT-Net & 0.04 & 0.07 & $+0.36$ & 0.08 & 0.06 & $-0.43$ \\
ObjectFormer & 0.14 & 0.22 & $+0.36$ & 0.19 & 0.15 & $-0.23$ \\
PSCC-Net & 0.13 & 0.16 & $+0.20$ & 0.17 & 0.14 & $-0.23$ \\
IML-ViT & 0.13 & 0.18 & $+0.27$ & 0.20 & 0.18 & $-0.08$ \\
TruFor & 0.08 & 0.14 & $+0.44$ & 0.11 & 0.12 & $+0.07$ \\
SIDA-7B & 0.01 & 0.03 & $+0.76$ & 0.01 & 0.03 & $+0.80$ \\
SparseViT & 0.08 & 0.12 & $+0.33$ & 0.12 & 0.09 & $-0.40$ \\
FMAE & 0.19 & 0.23 & $+0.17$ & 0.26 & 0.24 & $-0.11$ \\
ForgeryTalker & 0.13 & 0.21 & $+0.37$ & 0.11 & 0.17 & $+0.39$ \\
\bottomrule\end{tabular*}
\end{resulttable}
\end{minipage}
\end{table*}

\begin{figure*}[t]
\centering
\includegraphics[width=\textwidth]{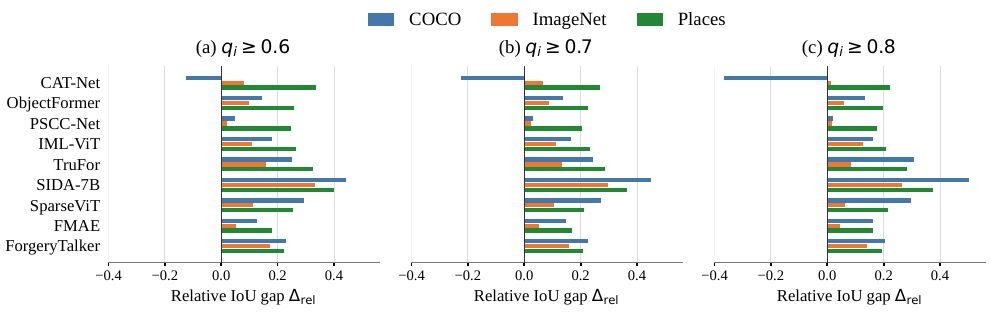}
\caption{Relative IoU gaps for nine localizers at pair-mask IoU thresholds of (a) 0.6, (b) 0.7, and (c) 0.8. Each model has one bar per dataset. Positive values indicate lower IoU on plausible edits.}
\label{fig:mask_overlap}
\end{figure*}

\subsection{Metrics and Evaluation}
\label{sec:protocol}\label{sec:metrics}
We test whether a localizer identifies plausible and anomalous edits of the same target equally well. \dataset{} is test-only: each of the nine public localizers receives one edited image and predicts a manipulation mask without access to the source or editing instruction. Each output is scored against its own reference mask. Predictions $B_i^z$ are binarized at 0.5 for both semantic conditions. For $J_i^z=\iou(B_i^z,M_i^z)$ over $N$ pairs, we report
\begin{equation}
\begin{aligned}
\bar J_{\pl}&=\frac1N\sum_{i=1}^N J_i^\pl,\quad
\bar J_{\an}=\frac1N\sum_{i=1}^N J_i^\an,\\
\Delta_{\rm rel}&=\frac{\bar J_{\an}-\bar J_{\pl}}{\bar J_{\an}}.
\end{aligned}
\label{eq:gap}
\end{equation}
A positive gap indicates lower plausible-edit IoU. Tables~\ref{tab:exp-coco}--\ref{tab:exp-places} report editor-specific and Avg results. Together, the two IoUs and the gap distinguish absolute accuracy from sensitivity to semantic plausibility.

\subsection{Localization across Models and Editors}
\label{sec:main_results}
We first ask whether the plausible-edit deficit is shared across localization approaches and editors. We evaluate nine public localizers spanning forensic cues, visual representations, and multimodal reasoning on the paired COCO, ImageNet, and Places test sets. Tables~\ref{tab:exp-coco}--\ref{tab:exp-places} show positive Avg gaps in 26 of 27 model--dataset comparisons, including all nine methods on ImageNet and Places. CAT-Net on COCO is the exception, with low IoUs in both conditions (0.054/0.051). The recurring sign across model families indicates that this evaluation axis exposes a weakness shared by otherwise different approaches.

On ImageNet, FMAE reaches plausible/anomalous IoUs of 0.273/0.288, compared with SparseViT's 0.090/0.103. Both retain positive gaps despite their different absolute accuracies. Editor-specific gaps also vary. On COCO, IML-ViT changes from $+0.279$ with Qwen to $-0.084$ with Flux.2, whereas ForgeryTalker has positive gaps for every editor across all three datasets. The paired scores capture these differences alongside overall accuracy. We also quantify the semantic-plausibility shortcut across the six intended violation types using the same pair IDs across localizers. After averaging IoUs equally over editors and datasets, relative gaps are positive in 53/54 model--type combinations, with CAT-Net on physical attributes (C1) the exception. The plausible-edit deficit thus recurs across different semantic conflicts. Appendix~\ref{app:category} provides the full condition-specific and dataset-specific statistics.

\subsection{Controls on Edit Magnitude and Region Size}
\label{sec:magnitude_controls}
\paragraph{Pixel-change distributions.}
We compare pixel-change distributions to test whether larger changes make anomalous edits easier to localize. For RGB vectors in $[0,1]^3$ and $z\in\{\pl,\an\}$, we define in-mask pixel change and paired change similarity as
\begin{align}
d_{\rm pix}(i,z)&=\frac{\sum_{p:M_i^z(p)=1}\|I_i^z(p)-I_i^{\real}(p)\|_1}{3|M_i^z|},
\label{eq:pixel-change}\\
s_{\rm pix}(i)&=\frac{\min(d_{\rm pix}(i,\pl),d_{\rm pix}(i,\an))}{\max(d_{\rm pix}(i,\pl),d_{\rm pix}(i,\an))}.
\label{eq:pixel-similarity}
\end{align}
Official-test means are 0.2636/0.2688 for plausible/anomalous edits (Table~\ref{tab:control-balance}). Across the release, 0.39\% of plausible and 0.26\% of anomalous outputs have $d_{\rm pix}<0.05$. Both conditions have similar overall change magnitudes. Appendix~\ref{app:pixel} gives the full distributions.

\paragraph{Localization with matched pixel-change magnitude.}
Based on these distributions, we remove pairs with a near-source member ($d_{\rm pix}<0.05$, N), then apply $s_{\rm pix}\geq0.8$ (S8) or $\geq0.9$ (S9) to test whether the semantic shortcut persists under matched pixel changes. S9 retains 28.4\% of test pairs and reduces mean absolute paired difference from 0.0782 to 0.0127 (Table~\ref{tab:control-balance}). Gaps remain positive in 25/27 comparisons (Table~\ref{tab:matched-localization}), except CAT-Net and PSCC-Net on COCO. The plausible-edit deficit persists after matching pixel-change magnitude. N and S8 results appear in Appendix~\ref{app:pixel}.

\paragraph{Localization with matched mask areas.}
We test whether differences in edited-region area account for the plausible-edit localization deficit. For an output of size $H_i\times W_i$, let $A_i^z=|M_i^z|/(H_iW_i)$ denote its mask-area fraction. We define paired area similarity as
\begin{equation}
s_{\rm area}(i)=\frac{\min(A_i^\pl,A_i^\an)}{\max(A_i^\pl,A_i^\an)}.
\label{eq:area-similarity}
\end{equation}
We retain pairs with $s_{\rm area}\geq0.9$. This leaves 14,511 official-test pairs (42.0\%) and reduces mean absolute paired area difference from 0.0579 to 0.0144. Table~\ref{tab:area-matched-localization} reports results for all nine localizers, with positive gaps in 26/27 model--dataset comparisons. CAT-Net on COCO is the exception. Plausible edits remain harder to localize at matched mask extent. Appendix~\ref{app:area_matched} provides area-balance statistics, size-stratified localization and F1 results.

\paragraph{Mask generation and pair overlap.}
We threshold the per-pixel mean absolute source--output RGB difference at 0.5, with RGB values in $[0,1]$, to obtain instruction-independent masks. With the prediction threshold fixed at 0.5, gaps remain positive in 26/27 model--dataset comparisons across nine localizers, except CAT-Net on COCO. The deficit persists with these masks, with full results in Appendix~\ref{app:diff_mask}. Figure~\ref{fig:mask_overlap} tests SIGMA pair-mask IoU thresholds of 0.6, 0.7, and 0.8. Each yields 26/27 positive gaps with the same exception, supporting the deficit under stricter spatial overlap.

\subsection{Human Validation}
\label{sec:human}
\paragraph{Mask quality and realized violations.}
Three experts review 500 English pairs covering all editors and violation types. They rate how closely each mask matches the visible edit on a scale from 1 (worst) to 5 (best), check whether the intended violation is visible in the anomalous output, and identify plausible outputs that are visually near-copies of the source. Plausible and anomalous masks receive mean scores of 4.586 and 4.576, with 90.6\% and 91.0\% rated at least 4, indicating similar mask quality across conditions. Experts confirm 372/500 violations (74.4\%) and find no visually near-copy plausible output. The Human Visual Assessment appendix (Appendix~\ref{app:human}) provides detailed statistics and further visual examples covering all six violation types, unrealized violations, and low-quality masks.

\paragraph{Localization on realized versus unrealized violations.}
Using these judgments, we split the 500 pairs into 372 realized and 128 unrealized cases, retaining each plausible--anomalous pairing. Table~\ref{tab:realization} shows positive gaps for all nine localizers on realized violations, compared with only SIDA-7B, ForgeryTalker, and TruFor on unrealized ones. For ObjectFormer, plausible/anomalous IoUs are $0.1419/0.2216$ (gap $+0.360$) on realized cases and $0.1874/0.1520$ (gap $-0.233$) on unrealized cases. When the intended anomaly is visible, all nine models localize the anomalous edit better than its plausible counterpart. This dependence on semantic abnormality further supports the semantic-plausibility shortcut in current localization models.

\subsection{Generalization across Domains and Evaluation Settings}
\label{sec:generalization}
\paragraph{Stylized images.}
To test whether the plausible-edit deficit extends beyond photographs, we apply the same paired construction and evaluation pipeline to cartoon images using Qwen-Image-Edit. The resulting 1,341 retained pairs satisfy the same pair-mask IoU requirement of at least 0.5. All nine localizers have positive relative gaps, ranging from 0.081 to 0.572. The deficit therefore also appears in this stylized domain. Appendix~\ref{app:stylized} provides the collection settings and full per-model scores.

\paragraph{Non-edited anomalous scenes.}
We examine false-positive localization on 500 non-edited anomalous photographs, whose unusual apparent scales or interactions arise from perspective, occlusion, and viewpoint. At threshold 0.5, mean false-positive pixel area ranges from 0.37\% to 11.99\% across localizers, while the fraction of images with at least one positive pixel ranges from 1.8\% to 100.0\%. These responses show that unusual content can be flagged even when no edited region exists. Appendix~\ref{app:unedited} provides collection and review details, prediction visualizations, and false-positive statistics at multiple thresholds.

\paragraph{Commercial localization service.}
We evaluate TextIn, a commercial image-manipulation detection service, on the same 500 audited English pairs as the nine public localizers. Returned regions are rasterized, thresholded at confidence 0.5, and scored against each output's SIGMA mask. TextIn obtains plausible/anomalous IoUs of 0.072/0.122 and a relative gap of 0.412. All ten systems have positive gaps on this shared cohort, extending the plausible-edit deficit to a commercial service. Appendix~\ref{app:textin} provides the service settings and full comparison with public checkpoints.

\section{Conclusion}
\rev{Manipulation localization should remain reliable when an edit is semantically plausible, yet existing benchmarks do not isolate this factor from differences in source content, editing models, and edited regions. We address this gap with \dataset{}, a paired benchmark that matches plausible and anomalous replacements of the same target while controlling the source, editor, language, generation settings, and spatial support. Its 47,747 bilingual triplets enable direct condition-specific evaluation, and experiments with nine public localizers reveal lower IoU on plausible edits in 26 of 27 model--dataset comparisons. This pattern remains under controls for edit magnitude, region size, evaluation metric, and mask construction. Together, the paired protocol, benchmark, and human audit identify the semantic-plausibility shortcut and provide a foundation for developing localizers that remain sensitive to edits even when the result looks ordinary. Future work can build on \dataset{} to design training objectives and representations that reduce this shortcut across broader edit operations and visual domains.}

\section{\rev{Limitations}}
\rev{\dataset{} establishes a controlled paired test across four editors and three photographic domains, with high and similar human-rated mask quality across semantic conditions. Several aspects can be strengthened further. Automatic mask estimation and a 74.4\% violation-realization rate leave room for more precise output verification and annotation. The current benchmark focuses on replacement edits and has limited coverage beyond photographs, which can be expanded to other edit operations, editors, and visual domains. The paired design controls major generation factors and measured edit properties, but it does not separate semantic plausibility from every editor-specific trace. Future work can address these directions through stronger validation, broader data coverage, and training studies that directly test and reduce the semantic-plausibility shortcut.}

\section*{LLM Use Statement}

In this work, we used generative AI tools to aid and polish writing: to edit the manuscript for readability, terminology consistency, and grammar. We have reviewed all AI-assisted text. The authors verified every claim, citation, and number against the source records and take responsibility for the final content of this work, including text produced with the aid of generative AI.

\label{main:lastpage}
{\small
\bibliographystyle{ieeenat_fullname}
\bibliography{references}
}

\clearpage

\clearpage
\appendix
\setcounter{table}{0}
\setcounter{figure}{0}
\setcounter{equation}{0}
\renewcommand{\thetable}{S\arabic{table}}
\renewcommand{\thefigure}{S\arabic{figure}}
\renewcommand{\theequation}{S\arabic{equation}}
\renewcommand{\theHtable}{S\arabic{table}}
\renewcommand{\theHfigure}{S\arabic{figure}}
\renewcommand{\theHequation}{S\arabic{equation}}
\raggedbottom
\section*{Appendix}
\makeatletter
\setlength{\@fptop}{0pt}
\setlength{\@fpsep}{12pt}
\setlength{\@dblfptop}{0pt}
\setlength{\@dblfpsep}{12pt}
\makeatother
\renewcommand{\topfraction}{0.9}
\renewcommand{\dbltopfraction}{0.9}
\renewcommand{\floatpagefraction}{0.85}
\renewcommand{\dblfloatpagefraction}{0.85}
\renewcommand{\textfraction}{0.1}
\setlength{\floatsep}{10pt plus 2pt minus 2pt}
\setlength{\textfloatsep}{10pt plus 2pt minus 2pt}
\setlength{\dblfloatsep}{10pt plus 2pt minus 2pt}
\setlength{\dbltextfloatsep}{10pt plus 2pt minus 2pt}
\setlength{\intextsep}{10pt plus 2pt minus 2pt}

\section{Pair Records}

\label{app:pair_record}
\paragraph{Annotated pair unit.}
Let $I_i^{\real}$ denote source photograph $i$. For semantic condition $z\in\{\pl,\an\}$, editor $\mathcal{G}_{e_i}$ applies instruction $P_i^z$ under shared controlled editing settings $\theta_i$:
\begin{equation}
I_i^z=\mathcal{G}_{e_i}(I_i^{\real},P_i^z;\theta_i).
\label{eq:generation}
\end{equation}
The two instructions replace the same target, while $\theta_i$ fixes the editor-side editing configuration. The resulting pair is the annotated unit
\begin{equation}
\begin{aligned}
u_i={}&(I_i^{\real},I_i^\pl,I_i^\an,M_i^\pl,M_i^\an,\\
       &\ P_i^\pl,P_i^\an,c_i,e_i),
\end{aligned}
\end{equation}
where $M_i^z$ is the mask of the realized edit on output $I_i^z$, $c_i$ is the \tax{} category sampled for pair $i$, and $e_i$ identifies the editor. Both edited outputs receive manipulation label $y=1$, with intended anomaly labels $a^\pl=0$ and $a^\an=1$. The source $I_i^{\real}$ is labeled $y=0$. The primary observable on $u_i$ is the \emph{plausibility-associated localization gap}: how much localization quality changes between its plausible and anomalous members.

\section{Stage A Prompt Templates and Validation}
\label{app:stage_a_prompts}

This appendix lists the English-facing Stage~A prompt templates and the deterministic validation rules used in \dataset{} construction.
Placeholders such as \lstinline|{objects_block}| and \lstinline|{anomaly_block}| are filled at runtime.
The production pipeline returns strict JSON with paired Chinese (\verb|_zh|) and English (\verb|_en|) fields.

\subsection{Perception module template}
\label{app:prompt_perception}

The perception module queries the source image with a fixed multimodal template to enumerate editable objects before pair generation.

\promptrole{System}
\begin{lstlisting}[style=PromptBox]
You are a rigorous assistant for constructing image-editing datasets for image manipulation localization research.
You must first identify visible objects in the image, and later generate paired edit instructions for the same object based on a replace operation.
Return strict JSON only. Do not output markdown, code fences, reasoning traces, or any text outside JSON.
The first character of your reply must be "{" and the last character must be "}".
\end{lstlisting}

\promptrole{User}
\begin{lstlisting}[style=PromptBox]
Carefully inspect the input image and list all clearly visible objects that can be edited independently.

Requirements:
- List only genuinely visible objects; do not invent off-screen or fully occluded objects.
- For each object, provide a short name, its spatial region, and a brief appearance description.
- Order objects from large to small, or from foreground to background.
- List at least 3 objects; if fewer are visible, list all of them.

Return JSON only, in the following format:
{
  "detected_objects": [
    {
      "object": "object",
      "region": "location of the object in the image",
      "description": "one-sentence description of the object appearance or state"
    }
  ]
}
\end{lstlisting}

\subsection{Prompt generation module template}
\label{app:prompt_pair}

Conditioned on the perception record, sampled category $c_i$, and source image, the prompt generation module selects one target and emits a bilingual plausible--anomalous replace pair with explicit semantic rationales and pair-control notes.

\promptrole{System}
\begin{lstlisting}[style=PromptBox]
You are a rigorous assistant for constructing image-editing datasets for image manipulation localization research.
Based on the already identified objects, generate both a normal (plausible) and an abnormal (anomalous) replace-style edit instruction for the same target object.
Fields ending with _zh must be written in Simplified Chinese; fields ending with _en must be accurate, natural English.
Return strict JSON only. Do not output markdown, code fences, reasoning traces, or any text outside JSON.
The first character of your reply must be "{" and the last character must be "}".
\end{lstlisting}

\promptrole{User (core)}
\begin{lstlisting}[style=PromptBox]
Below is the list of objects detected in the image:
{objects_block}

Select exactly 1 object from the list that is most suitable for a local replace edit, and generate a bilingual pair of edit instructions.

Editing requirements:
1. Both the normal and abnormal instructions must be replace-style: Chinese must contain the release replace keyword, and English must use "Replace ...".
2. normal_prompt_*: after replacement, the result must be semantically plausible and consistent with common sense and physical laws.
3. abnormal_prompt_*: the replacement target must still be visually renderable, but must introduce a logical contradiction, physical implausibility, or common-sense violation.
4. Both instructions must edit the same object in the same region, and should preserve scale, viewpoint, lighting, and style as much as possible.
5. Edit only one target object; do not modify multiple unrelated objects or redraw the whole image.
6. Avoid violent, sexual, hateful, or gory content.
7. The English version must faithfully convey the Chinese version and must not introduce any new edit target absent from the Chinese version.
{anomaly_block}

Return JSON only, with the following fields:
{
  "target_object_zh": "...",
  "target_object_en": "...",
  "target_region_zh": "...",
  "target_region_en": "...",
  "target_object_description_zh": "...",
  "target_object_description_en": "...",
  "anomaly_type": {anomaly_type_field},
  "normal_prompt_zh": "...",
  "normal_prompt_en": "...",
  "abnormal_prompt_zh": "...",
  "abnormal_prompt_en": "...",
  "normal_semantic_rationale_zh": "...",
  "normal_semantic_rationale_en": "...",
  "abnormal_semantic_rationale_zh": "...",
  "abnormal_semantic_rationale_en": "...",
  "pair_control_notes_zh": "...",
  "pair_control_notes_en": "..."
}
\end{lstlisting}

When a \tax{} category is sampled, \lstinline|{anomaly_block}| expands to the category id, bilingual name, definition, in-category examples, and guardrails.
In auto mode it instead instructs the model to choose an anomaly type from the selected object and scene.

\subsection{Prompt validation module (SPPV)}
\label{app:prompt_validation}

Stage~A uses deterministic structured-prompt validation after parsing. Required-field and replacement-form checks operate on the instruction record. The taxonomy and semantic rationales specify the intended contrast and their presence does not verify how the editor realizes it. 

\paragraph{Complete grounding.}
Every bilingual schema field must be present and non-empty: target object, target region, pre-edit description, \texttt{anomaly\_type}, normal/abnormal prompts, semantic rationales, and pair-control notes (Chinese and English counterparts).
\rev{These fields record the shared target and region identified during perception.}

\paragraph{Edit-form consistency.}
\rev{Replacement wording keeps the requested edit operation fixed across conditions.}

\paragraph{Taxonomy-aligned contrast.}
\rev{The schema records the sampled category $c_i$ and paired semantic rationales. These fields describe the intended conflict and compatible alternative for the shared target. The pseudocode below describes the documented structural criteria.}

\begin{lstlisting}[style=PromptBox]
SPPV(pair) -> ACCEPT iff all checks pass; otherwise REJECT

Complete grounding:
  all required bilingual fields present and non-empty

Edit-form consistency:
  release replace keyword in normal_prompt_zh and abnormal_prompt_zh
  "replace" in normal_prompt_en and abnormal_prompt_en (case-insensitive)

Taxonomy-aligned contrast:
  anomaly_type == sampled SSTOA category c_i  (taxonomy mode)
  normal_semantic_rationale_* and abnormal_semantic_rationale_* non-empty
\end{lstlisting}

\section{Construction Controls and Prompt Validation}
\label{app:editor_design}
For a given source photograph and specific language, all four editors receive exactly the same plausible and anomalous prompts. Within each editor, the two outputs share its generation configuration. Across languages, the source, target, intended category, and editor are shared, while each language produces its own images and masks. This design broadens instruction coverage. 

The public editors support instruction-guided replacement and differ in backbone scale and conditioning: Qwen-Image-Edit-2511 uses a 20B double-stream multimodal diffusion transformer with Qwen2.5-VL conditioning. FireRed-Image-Edit-1.1 builds on the Qwen-Image foundation with unified 3D rotary position embeddings for reference and target tokens. LongCat-Image-Edit uses a 6B backbone with double- and single-stream blocks. Flux.2-dev uses a 32B rectified-flow transformer~\citep{wu2025qwenimage,qiao2026firered,longcat2025image,blackforest2025flux2}. Each editor retains its own generation configuration.

\begingroup
\captionsetup{hypcap=false}
\begin{table*}[p]\centering\footnotesize
\caption{\rev{Localization results by violation type across three datasets. }}\label{tab:category-overall}
\setlength{\tabcolsep}{1.2pt}
\begin{tabular*}{\textwidth}{@{\extracolsep{\fill}}l*{18}{r}@{}}\toprule
 & \multicolumn{3}{c}{C1} & \multicolumn{3}{c}{C2} & \multicolumn{3}{c}{C3} & \multicolumn{3}{c}{C4} & \multicolumn{3}{c}{C5} & \multicolumn{3}{c}{C6} \\
\cmidrule(lr){2-4}\cmidrule(lr){5-7}\cmidrule(lr){8-10}\cmidrule(lr){11-13}\cmidrule(lr){14-16}\cmidrule(lr){17-19}
Method & $J_\pl$ & $J_\an$ & $\Delta_{\rm rel}$ & $J_\pl$ & $J_\an$ & $\Delta_{\rm rel}$ & $J_\pl$ & $J_\an$ & $\Delta_{\rm rel}$ & $J_\pl$ & $J_\an$ & $\Delta_{\rm rel}$ & $J_\pl$ & $J_\an$ & $\Delta_{\rm rel}$ & $J_\pl$ & $J_\an$ & $\Delta_{\rm rel}$ \\\midrule
CAT-Net & 0.058 & 0.052 & -0.118 & 0.056 & 0.075 & 0.250 & 0.072 & 0.084 & 0.138 & 0.046 & 0.070 & 0.340 & 0.048 & 0.066 & 0.278 & 0.053 & 0.065 & 0.183 \\
ObjectFormer & 0.155 & 0.174 & 0.107 & 0.172 & 0.213 & 0.193 & 0.180 & 0.215 & 0.162 & 0.180 & 0.211 & 0.148 & 0.155 & 0.176 & 0.120 & 0.174 & 0.264 & 0.340 \\
PSCC-Net & 0.119 & 0.119 & 0.001 & 0.146 & 0.169 & 0.136 & 0.143 & 0.167 & 0.147 & 0.131 & 0.161 & 0.189 & 0.142 & 0.147 & 0.039 & 0.144 & 0.175 & 0.178 \\
IML-ViT & 0.141 & 0.156 & 0.101 & 0.149 & 0.203 & 0.267 & 0.167 & 0.198 & 0.156 & 0.144 & 0.184 & 0.218 & 0.124 & 0.163 & 0.238 & 0.131 & 0.157 & 0.168 \\
TruFor & 0.081 & 0.104 & 0.221 & 0.088 & 0.115 & 0.238 & 0.112 & 0.150 & 0.255 & 0.072 & 0.120 & 0.400 & 0.086 & 0.110 & 0.223 & 0.072 & 0.098 & 0.269 \\
SIDA-7B & 0.027 & 0.030 & 0.107 & 0.015 & 0.038 & 0.605 & 0.028 & 0.062 & 0.540 & 0.031 & 0.048 & 0.354 & 0.021 & 0.034 & 0.375 & 0.029 & 0.040 & 0.270 \\
SparseViT & 0.084 & 0.097 & 0.138 & 0.101 & 0.124 & 0.189 & 0.108 & 0.154 & 0.296 & 0.079 & 0.115 & 0.317 & 0.094 & 0.110 & 0.139 & 0.077 & 0.108 & 0.284 \\
FMAE & 0.204 & 0.243 & 0.163 & 0.214 & 0.239 & 0.106 & 0.243 & 0.246 & 0.011 & 0.225 & 0.267 & 0.155 & 0.207 & 0.212 & 0.025 & 0.191 & 0.236 & 0.189 \\
ForgeryTalker & 0.139 & 0.169 & 0.177 & 0.152 & 0.185 & 0.180 & 0.154 & 0.184 & 0.165 & 0.173 & 0.245 & 0.291 & 0.155 & 0.200 & 0.226 & 0.157 & 0.200 & 0.213 \\
\bottomrule\end{tabular*}
\par\medskip
\caption{\rev{Localization resultsby violation type on COCO.}}\label{tab:category-coco}
\setlength{\tabcolsep}{1.2pt}
\begin{tabular*}{\textwidth}{@{\extracolsep{\fill}}l*{18}{r}@{}}\toprule
 & \multicolumn{3}{c}{C1} & \multicolumn{3}{c}{C2} & \multicolumn{3}{c}{C3} & \multicolumn{3}{c}{C4} & \multicolumn{3}{c}{C5} & \multicolumn{3}{c}{C6} \\
\cmidrule(lr){2-4}\cmidrule(lr){5-7}\cmidrule(lr){8-10}\cmidrule(lr){11-13}\cmidrule(lr){14-16}\cmidrule(lr){17-19}
Method & $J_\pl$ & $J_\an$ & $\Delta_{\rm rel}$ & $J_\pl$ & $J_\an$ & $\Delta_{\rm rel}$ & $J_\pl$ & $J_\an$ & $\Delta_{\rm rel}$ & $J_\pl$ & $J_\an$ & $\Delta_{\rm rel}$ & $J_\pl$ & $J_\an$ & $\Delta_{\rm rel}$ & $J_\pl$ & $J_\an$ & $\Delta_{\rm rel}$ \\\midrule
CAT-Net & 0.054 & 0.026 & -1.041 & 0.068 & 0.098 & 0.305 & 0.078 & 0.062 & -0.260 & 0.029 & 0.037 & 0.220 & 0.042 & 0.045 & 0.072 & 0.048 & 0.054 & 0.125 \\
ObjectFormer & 0.144 & 0.172 & 0.164 & 0.213 & 0.244 & 0.130 & 0.215 & 0.244 & 0.115 & 0.219 & 0.224 & 0.024 & 0.127 & 0.162 & 0.216 & 0.225 & 0.313 & 0.282 \\
PSCC-Net & 0.109 & 0.110 & 0.008 & 0.168 & 0.199 & 0.155 & 0.135 & 0.151 & 0.107 & 0.128 & 0.145 & 0.114 & 0.140 & 0.103 & -0.361 & 0.165 & 0.196 & 0.157 \\
IML-ViT & 0.114 & 0.131 & 0.135 & 0.178 & 0.259 & 0.316 & 0.186 & 0.218 & 0.145 & 0.158 & 0.161 & 0.017 & 0.095 & 0.140 & 0.326 & 0.138 & 0.133 & -0.039 \\
TruFor & 0.046 & 0.063 & 0.272 & 0.099 & 0.123 & 0.195 & 0.114 & 0.149 & 0.234 & 0.026 & 0.073 & 0.643 & 0.091 & 0.094 & 0.033 & 0.054 & 0.074 & 0.269 \\
SIDA-7B & 0.027 & 0.031 & 0.117 & 0.004 & 0.030 & 0.867 & 0.038 & 0.082 & 0.535 & 0.033 & 0.056 & 0.415 & 0.021 & 0.035 & 0.405 & 0.043 & 0.037 & -0.145 \\
SparseViT & 0.050 & 0.079 & 0.374 & 0.118 & 0.153 & 0.229 & 0.087 & 0.146 & 0.409 & 0.045 & 0.076 & 0.406 & 0.091 & 0.087 & -0.042 & 0.068 & 0.080 & 0.147 \\
FMAE & 0.147 & 0.205 & 0.280 & 0.202 & 0.240 & 0.159 & 0.237 & 0.207 & -0.144 & 0.196 & 0.218 & 0.103 & 0.152 & 0.130 & -0.164 & 0.148 & 0.192 & 0.229 \\
ForgeryTalker & 0.110 & 0.152 & 0.276 & 0.108 & 0.129 & 0.162 & 0.139 & 0.180 & 0.228 & 0.174 & 0.222 & 0.213 & 0.149 & 0.176 & 0.153 & 0.126 & 0.160 & 0.210 \\
\bottomrule\end{tabular*}
\par\medskip
\caption{\rev{Localization resultsby violation type on ImageNet.}}\label{tab:category-imagenet}
\setlength{\tabcolsep}{1.2pt}
\begin{tabular*}{\textwidth}{@{\extracolsep{\fill}}l*{18}{r}@{}}\toprule
 & \multicolumn{3}{c}{C1} & \multicolumn{3}{c}{C2} & \multicolumn{3}{c}{C3} & \multicolumn{3}{c}{C4} & \multicolumn{3}{c}{C5} & \multicolumn{3}{c}{C6} \\
\cmidrule(lr){2-4}\cmidrule(lr){5-7}\cmidrule(lr){8-10}\cmidrule(lr){11-13}\cmidrule(lr){14-16}\cmidrule(lr){17-19}
Method & $J_\pl$ & $J_\an$ & $\Delta_{\rm rel}$ & $J_\pl$ & $J_\an$ & $\Delta_{\rm rel}$ & $J_\pl$ & $J_\an$ & $\Delta_{\rm rel}$ & $J_\pl$ & $J_\an$ & $\Delta_{\rm rel}$ & $J_\pl$ & $J_\an$ & $\Delta_{\rm rel}$ & $J_\pl$ & $J_\an$ & $\Delta_{\rm rel}$ \\\midrule
CAT-Net & 0.061 & 0.064 & 0.055 & 0.054 & 0.059 & 0.074 & 0.070 & 0.071 & 0.012 & 0.058 & 0.072 & 0.188 & 0.054 & 0.062 & 0.130 & 0.058 & 0.068 & 0.151 \\
ObjectFormer & 0.175 & 0.186 & 0.060 & 0.166 & 0.192 & 0.138 & 0.172 & 0.176 & 0.021 & 0.175 & 0.203 & 0.140 & 0.190 & 0.181 & -0.051 & 0.164 & 0.244 & 0.327 \\
PSCC-Net & 0.140 & 0.133 & -0.052 & 0.165 & 0.159 & -0.035 & 0.166 & 0.169 & 0.018 & 0.161 & 0.179 & 0.097 & 0.170 & 0.178 & 0.041 & 0.157 & 0.177 & 0.117 \\
IML-ViT & 0.166 & 0.184 & 0.098 & 0.152 & 0.169 & 0.100 & 0.169 & 0.174 & 0.027 & 0.148 & 0.186 & 0.206 & 0.151 & 0.166 & 0.088 & 0.139 & 0.177 & 0.214 \\
TruFor & 0.086 & 0.103 & 0.173 & 0.076 & 0.086 & 0.120 & 0.091 & 0.097 & 0.064 & 0.079 & 0.118 & 0.329 & 0.070 & 0.080 & 0.130 & 0.073 & 0.092 & 0.205 \\
SIDA-7B & 0.022 & 0.021 & -0.025 & 0.024 & 0.037 & 0.335 & 0.021 & 0.041 & 0.498 & 0.026 & 0.044 & 0.394 & 0.016 & 0.025 & 0.352 & 0.022 & 0.038 & 0.414 \\
SparseViT & 0.091 & 0.096 & 0.054 & 0.083 & 0.091 & 0.085 & 0.103 & 0.113 & 0.080 & 0.094 & 0.118 & 0.198 & 0.087 & 0.089 & 0.028 & 0.078 & 0.113 & 0.305 \\
FMAE & 0.268 & 0.313 & 0.144 & 0.259 & 0.262 & 0.010 & 0.283 & 0.264 & -0.070 & 0.294 & 0.319 & 0.079 & 0.272 & 0.264 & -0.030 & 0.265 & 0.307 & 0.136 \\
ForgeryTalker & 0.182 & 0.206 & 0.115 & 0.206 & 0.241 & 0.142 & 0.196 & 0.219 & 0.106 & 0.205 & 0.291 & 0.295 & 0.182 & 0.236 & 0.230 & 0.198 & 0.260 & 0.238 \\
\bottomrule\end{tabular*}
\par\medskip
\caption{\rev{Localization results by violation type on Places.}}\label{tab:category-places}
\setlength{\tabcolsep}{1.2pt}
\begin{tabular*}{\textwidth}{@{\extracolsep{\fill}}l*{18}{r}@{}}\toprule
 & \multicolumn{3}{c}{C1} & \multicolumn{3}{c}{C2} & \multicolumn{3}{c}{C3} & \multicolumn{3}{c}{C4} & \multicolumn{3}{c}{C5} & \multicolumn{3}{c}{C6} \\
\cmidrule(lr){2-4}\cmidrule(lr){5-7}\cmidrule(lr){8-10}\cmidrule(lr){11-13}\cmidrule(lr){14-16}\cmidrule(lr){17-19}
Method & $J_\pl$ & $J_\an$ & $\Delta_{\rm rel}$ & $J_\pl$ & $J_\an$ & $\Delta_{\rm rel}$ & $J_\pl$ & $J_\an$ & $\Delta_{\rm rel}$ & $J_\pl$ & $J_\an$ & $\Delta_{\rm rel}$ & $J_\pl$ & $J_\an$ & $\Delta_{\rm rel}$ & $J_\pl$ & $J_\an$ & $\Delta_{\rm rel}$ \\\midrule
CAT-Net & 0.060 & 0.066 & 0.084 & 0.046 & 0.068 & 0.324 & 0.069 & 0.119 & 0.419 & 0.051 & 0.100 & 0.493 & 0.047 & 0.091 & 0.480 & 0.053 & 0.072 & 0.257 \\
ObjectFormer & 0.147 & 0.164 & 0.100 & 0.137 & 0.202 & 0.321 & 0.153 & 0.225 & 0.322 & 0.145 & 0.205 & 0.292 & 0.149 & 0.186 & 0.201 & 0.134 & 0.236 & 0.431 \\
PSCC-Net & 0.108 & 0.114 & 0.055 & 0.105 & 0.148 & 0.294 & 0.127 & 0.182 & 0.300 & 0.103 & 0.161 & 0.360 & 0.115 & 0.162 & 0.290 & 0.110 & 0.152 & 0.275 \\
IML-ViT & 0.142 & 0.154 & 0.077 & 0.116 & 0.180 & 0.354 & 0.146 & 0.203 & 0.278 & 0.126 & 0.206 & 0.385 & 0.126 & 0.182 & 0.307 & 0.115 & 0.161 & 0.287 \\
TruFor & 0.111 & 0.145 & 0.233 & 0.089 & 0.137 & 0.350 & 0.131 & 0.204 & 0.361 & 0.111 & 0.169 & 0.346 & 0.096 & 0.157 & 0.386 & 0.088 & 0.129 & 0.314 \\
SIDA-7B & 0.030 & 0.037 & 0.176 & 0.017 & 0.048 & 0.648 & 0.026 & 0.061 & 0.575 & 0.034 & 0.045 & 0.238 & 0.026 & 0.041 & 0.363 & 0.022 & 0.044 & 0.497 \\
SparseViT & 0.111 & 0.116 & 0.046 & 0.101 & 0.129 & 0.214 & 0.135 & 0.202 & 0.334 & 0.097 & 0.153 & 0.364 & 0.106 & 0.153 & 0.306 & 0.085 & 0.130 & 0.350 \\
FMAE & 0.196 & 0.213 & 0.077 & 0.181 & 0.216 & 0.162 & 0.209 & 0.266 & 0.212 & 0.186 & 0.263 & 0.291 & 0.197 & 0.242 & 0.187 & 0.161 & 0.209 & 0.229 \\
ForgeryTalker & 0.125 & 0.149 & 0.160 & 0.140 & 0.185 & 0.241 & 0.126 & 0.152 & 0.175 & 0.141 & 0.221 & 0.363 & 0.134 & 0.189 & 0.288 & 0.147 & 0.180 & 0.181 \\
\bottomrule\end{tabular*}
\end{table*}
\endgroup

\section{Localization by Intended Violation Type}
\label{app:category}

Aggregate localization scores can conceal differences among violation types. We compare intended categories and source datasets to identify where the plausible-edit deficit appears. Table~\ref{tab:category-overall} summarizes the six intended types across datasets, while Tables~\ref{tab:category-coco}--\ref{tab:category-places} report each dataset separately. 

\paragraph{Coverage across violation types.}
The overall gap is positive in 53/54 model--category combinations (Table~\ref{tab:category-overall}). CAT-Net on physical attribute violations (C1) is the exception. Each of the six intended types has positive gaps for at least eight of the nine methods. The plausible-edit deficit therefore spans the full set of semantic conflicts represented in the benchmark.

\paragraph{Variation across source datasets.}
Tables~\ref{tab:category-coco}--\ref{tab:category-places} show positive gaps in 46/54 COCO, 48/54 ImageNet, and 54/54 Places comparisons, totaling 148/162. Spatial-relation violations have positive gaps in all 27 model--dataset combinations. The pattern occurs in every source domain, with the most consistent sign pattern on Places, while the dataset-specific tables identify the categories and methods where the gap reverses.

\section{Pixel-Change Distributions and Controlled Localization}
\label{app:pixel}
\begin{figure*}[tp]
\centering
\includegraphics[width=\textwidth]{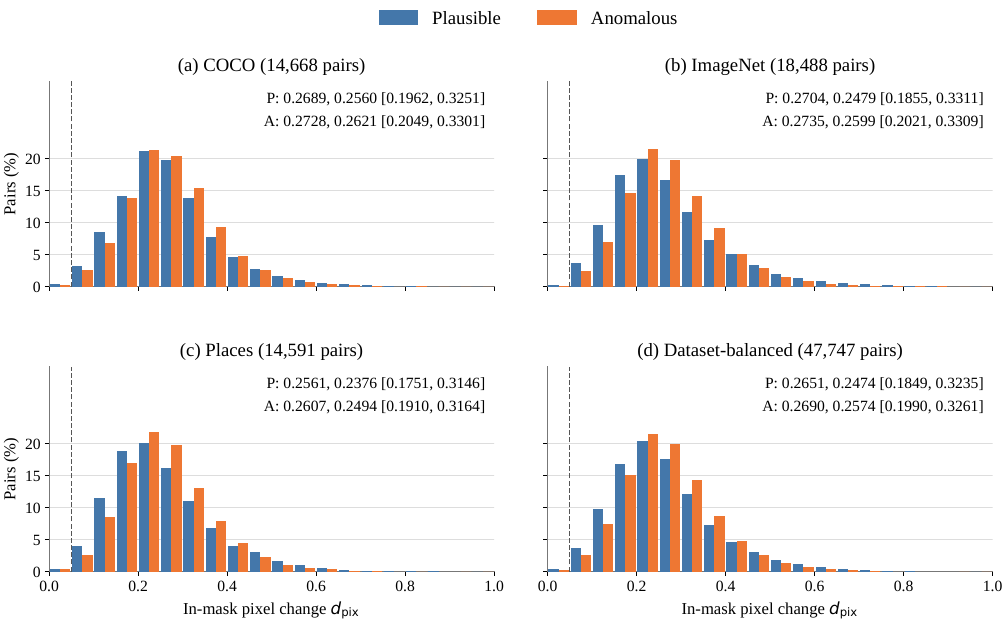}
\caption{In-mask pixel-change distributions for the 47,747 release pairs, with bins of width 0.05. P and A denote plausible and anomalous edits. Each panel reports the mean and median [Q25, Q75] for both conditions. The dashed line marks the near-copy threshold of 0.05.}
\label{fig:pixel-distributions}
\end{figure*}

Pixel change and paired change similarity follow main-paper Equations~\ref{eq:pixel-change} and~\ref{eq:pixel-similarity}. We examine whether plausible edits are near-copies, whether they systematically involve smaller pixel changes, and whether the localization gap persists after matching edit magnitude. Figure~\ref{fig:pixel-distributions} describes the change distributions. Tables~\ref{tab:ordering} and~\ref{tab:paired-difference} quantify within-pair differences, while Tables~\ref{tab:pixel-coco}--\ref{tab:pixel-places} report localization under progressively stricter magnitude controls.

\paragraph{Near-copy prevalence and change distributions.}
Using the measure in main-paper Equation~\ref{eq:pixel-change}, near-source outputs have $d_{\rm pix}<0.05$. Across all 47,747 release pairs, only 0.39\% of plausible edits and 0.26\% of anomalous edits fall below this threshold (Figure~\ref{fig:pixel-distributions}). Mean changes are 0.2651/0.2690 and medians are 0.2474/0.2574 (Figure~\ref{fig:pixel-distributions}). Near-copies are therefore uncommon under this criterion, and both conditions occupy similar magnitude ranges.

\paragraph{Direction and magnitude of paired differences.}
Plausible change is smaller in 52.92\% of pairs and at least as large in 47.08\%, with ties included in the latter group (Table~\ref{tab:ordering}). The anomalous-minus-plausible mean difference is +0.0039, with a 95\% confidence interval of +0.0015 to +0.0063 (Table~\ref{tab:paired-difference}). The interval resamples source-image groups while keeping their editor and language variants together. Both directions of change are common, with a small average shift toward larger anomalous changes. The localization controls below compare pairs with increasingly similar magnitudes.

\begin{table}[htbp]\centering\small

\caption{\rev{Paired change ordering, with ties assigned to the final column.}}\label{tab:ordering}

\setlength{\tabcolsep}{4pt}\renewcommand{\arraystretch}{1.12}

\begin{lrbox}{\revisiontablebox}

\begin{tabular}{@{}lcc@{}}\toprule

Dataset & $d_{\rm pix}^{\pl}<d_{\rm pix}^{\an}$ & $d_{\rm pix}^{\pl}\geq d_{\rm pix}^{\an}$ \\\midrule

COCO & 51.81\% & 48.19\% \\

ImageNet & 53.95\% & 46.05\% \\

Places & 52.99\% & 47.01\% \\

Mean across datasets & 52.92\% & 47.08\% \\

\bottomrule\end{tabular}\end{lrbox}

\ifdim\wd\revisiontablebox>\linewidth\resizebox{\linewidth}{!}{\usebox{\revisiontablebox}}\else\usebox{\revisiontablebox}\fi

\end{table}

\begin{table}[htbp]\centering\footnotesize

\caption{\rev{Paired anomalous-minus-plausible change. Intervals resample source-image groups.}}\label{tab:paired-difference}

\setlength{\tabcolsep}{2pt}\renewcommand{\arraystretch}{1.12}

\begin{lrbox}{\revisiontablebox}

\begin{tabular}{@{}lcc@{}}\toprule

Dataset & Mean [95\% CI] & Median [Q25, Q75] \\\midrule

COCO & \shortstack{+0.0039\\{[-0.0003, +0.0084]}} & \shortstack{+0.0026\\{[-0.0426, +0.0541]}} \\

ImageNet & \shortstack{+0.0032\\{[-0.0009, +0.0072]}} & \shortstack{+0.0059\\{[-0.0459, +0.0579]}} \\

Places & \shortstack{+0.0047\\{[+0.0002, +0.0091]}} & \shortstack{+0.0043\\{[-0.0469, +0.0573]}} \\

Dataset-balanced & \shortstack{+0.0039\\{[+0.0015, +0.0063]}} & \shortstack{+0.0043\\{[-0.0451, +0.0565]}} \\

\bottomrule\end{tabular}\end{lrbox}

\ifdim\wd\revisiontablebox>\linewidth\resizebox{\linewidth}{!}{\usebox{\revisiontablebox}}\else\usebox{\revisiontablebox}\fi

\end{table}

\paragraph{Localization after magnitude matching.}
We use the same official-test pair IDs for all nine localizers. N removes a pair if either member has $d_{\rm pix}<0.05$, while S8 and S9 additionally require the similarity in main-paper Equation~\ref{eq:pixel-similarity} to be at least 0.8 and 0.9. These subsets retain 99.5\%, 51.3\%, and 28.4\% of the original official-test pairs. At S9, mean changes are 0.2585/0.2591 and the mean absolute paired difference decreases from 0.0782 to 0.0127 (Table~\ref{tab:control-balance}). These statistics describe the official-test cohort, separately from the full-release distributions in Figure~\ref{fig:pixel-distributions}. Tables~\ref{tab:pixel-coco}--\ref{tab:pixel-places} show positive gaps in 26/27 comparisons for N and S8 and 25/27 for S9. The S8 exception is PSCC-Net on COCO, and the S9 exceptions are CAT-Net and PSCC-Net on COCO. Every method retains a positive overall gap after averaging the condition-specific IoUs equally across datasets. The plausible-edit deficit thus persists when measured pixel change is closely matched.

\begin{table*}[tp]\centering\footnotesize
\caption{\rev{Localization under pixel-change controls on COCO. Base gives the main-table values. N requires both changes to be at least 0.05. S8 and S9 additionally require similarity of at least 0.8 and 0.9.}}\label{tab:pixel-coco}
\setlength{\tabcolsep}{2.5pt}\renewcommand{\arraystretch}{1.04}
\begin{tabular*}{\textwidth}{@{\extracolsep{\fill}}l*{12}{r}@{}}\toprule
 & \multicolumn{3}{c}{Base} & \multicolumn{3}{c}{N} & \multicolumn{3}{c}{S8} & \multicolumn{3}{c}{S9} \\
\cmidrule(lr){2-4}\cmidrule(lr){5-7}\cmidrule(lr){8-10}\cmidrule(lr){11-13}
Method & $J_\pl$ & $J_\an$ & $\Delta_{\rm rel}$ & $J_\pl$ & $J_\an$ & $\Delta_{\rm rel}$ & $J_\pl$ & $J_\an$ & $\Delta_{\rm rel}$ & $J_\pl$ & $J_\an$ & $\Delta_{\rm rel}$ \\\midrule
CAT-Net & 0.0540 & 0.0510 & -0.059 & 0.0541 & 0.0507 & -0.066 & 0.0499 & 0.0511 & +0.023 & 0.0547 & 0.0530 & -0.033 \\
ObjectFormer & 0.1860 & 0.2210 & +0.159 & 0.1867 & 0.2223 & +0.160 & 0.1786 & 0.2110 & +0.153 & 0.1712 & 0.2035 & +0.158 \\
PSCC-Net & 0.1370 & 0.1460 & +0.062 & 0.1379 & 0.1462 & +0.057 & 0.1462 & 0.1423 & -0.027 & 0.1463 & 0.1322 & -0.107 \\
IML-ViT & 0.1420 & 0.1700 & +0.165 & 0.1426 & 0.1703 & +0.163 & 0.1464 & 0.1739 & +0.158 & 0.1354 & 0.1686 & +0.197 \\
TruFor & 0.0700 & 0.0940 & +0.255 & 0.0707 & 0.0947 & +0.254 & 0.0747 & 0.0937 & +0.203 & 0.0822 & 0.0975 & +0.157 \\
SIDA-7B & 0.0280 & 0.0450 & +0.378 & 0.0285 & 0.0453 & +0.372 & 0.0246 & 0.0436 & +0.434 & 0.0228 & 0.0366 & +0.378 \\
SparseViT & 0.0740 & 0.1020 & +0.275 & 0.0739 & 0.1023 & +0.278 & 0.0871 & 0.1067 & +0.184 & 0.0864 & 0.1076 & +0.197 \\
FMAE & 0.1780 & 0.1990 & +0.106 & 0.1789 & 0.1995 & +0.103 & 0.1900 & 0.2043 & +0.070 & 0.1781 & 0.1997 & +0.108 \\
ForgeryTalker & 0.1320 & 0.1680 & +0.214 & 0.1325 & 0.1687 & +0.214 & 0.1395 & 0.1713 & +0.186 & 0.1384 & 0.1679 & +0.176 \\
\bottomrule\end{tabular*}
\end{table*}

\begin{table*}[tp]\centering\footnotesize
\caption{\rev{Localization under pixel-change controls on ImageNet. Base gives the main-table values. N requires both changes to be at least 0.05. S8 and S9 additionally require similarity of at least 0.8 and 0.9.}}\label{tab:pixel-imagenet}
\setlength{\tabcolsep}{2.5pt}\renewcommand{\arraystretch}{1.04}
\begin{tabular*}{\textwidth}{@{\extracolsep{\fill}}l*{12}{r}@{}}\toprule
 & \multicolumn{3}{c}{Base} & \multicolumn{3}{c}{N} & \multicolumn{3}{c}{S8} & \multicolumn{3}{c}{S9} \\
\cmidrule(lr){2-4}\cmidrule(lr){5-7}\cmidrule(lr){8-10}\cmidrule(lr){11-13}
Method & $J_\pl$ & $J_\an$ & $\Delta_{\rm rel}$ & $J_\pl$ & $J_\an$ & $\Delta_{\rm rel}$ & $J_\pl$ & $J_\an$ & $\Delta_{\rm rel}$ & $J_\pl$ & $J_\an$ & $\Delta_{\rm rel}$ \\\midrule
CAT-Net & 0.0590 & 0.0660 & +0.106 & 0.0595 & 0.0657 & +0.095 & 0.0593 & 0.0685 & +0.135 & 0.0599 & 0.0701 & +0.145 \\
ObjectFormer & 0.1740 & 0.1950 & +0.108 & 0.1740 & 0.1954 & +0.110 & 0.1643 & 0.1831 & +0.103 & 0.1630 & 0.1818 & +0.103 \\
PSCC-Net & 0.1590 & 0.1630 & +0.025 & 0.1588 & 0.1634 & +0.028 & 0.1621 & 0.1658 & +0.022 & 0.1626 & 0.1662 & +0.021 \\
IML-ViT & 0.1560 & 0.1760 & +0.114 & 0.1560 & 0.1766 & +0.117 & 0.1514 & 0.1716 & +0.118 & 0.1523 & 0.1702 & +0.105 \\
TruFor & 0.0800 & 0.0960 & +0.167 & 0.0800 & 0.0964 & +0.171 & 0.0782 & 0.0979 & +0.201 & 0.0802 & 0.0970 & +0.174 \\
SIDA-7B & 0.0220 & 0.0340 & +0.353 & 0.0220 & 0.0336 & +0.344 & 0.0223 & 0.0339 & +0.344 & 0.0227 & 0.0337 & +0.326 \\
SparseViT & 0.0900 & 0.1030 & +0.126 & 0.0900 & 0.1025 & +0.122 & 0.0873 & 0.1053 & +0.170 & 0.0920 & 0.1070 & +0.141 \\
FMAE & 0.2730 & 0.2880 & +0.052 & 0.2732 & 0.2888 & +0.054 & 0.2660 & 0.2820 & +0.057 & 0.2639 & 0.2800 & +0.057 \\
ForgeryTalker & 0.1940 & 0.2380 & +0.185 & 0.1942 & 0.2382 & +0.185 & 0.1869 & 0.2306 & +0.190 & 0.1888 & 0.2308 & +0.182 \\
\bottomrule\end{tabular*}
\end{table*}

\begin{table*}[tp]\centering\footnotesize
\caption{\rev{Localization under pixel-change controls on Places. Base gives the main-table values. N requires both changes to be at least 0.05. S8 and S9 additionally require similarity of at least 0.8 and 0.9.}}\label{tab:pixel-places}
\setlength{\tabcolsep}{2.5pt}\renewcommand{\arraystretch}{1.04}
\begin{tabular*}{\textwidth}{@{\extracolsep{\fill}}l*{12}{r}@{}}\toprule
 & \multicolumn{3}{c}{Base} & \multicolumn{3}{c}{N} & \multicolumn{3}{c}{S8} & \multicolumn{3}{c}{S9} \\
\cmidrule(lr){2-4}\cmidrule(lr){5-7}\cmidrule(lr){8-10}\cmidrule(lr){11-13}
Method & $J_\pl$ & $J_\an$ & $\Delta_{\rm rel}$ & $J_\pl$ & $J_\an$ & $\Delta_{\rm rel}$ & $J_\pl$ & $J_\an$ & $\Delta_{\rm rel}$ & $J_\pl$ & $J_\an$ & $\Delta_{\rm rel}$ \\\midrule
CAT-Net & 0.0560 & 0.0840 & +0.333 & 0.0552 & 0.0846 & +0.347 & 0.0558 & 0.0824 & +0.323 & 0.0528 & 0.0804 & +0.344 \\
ObjectFormer & 0.1450 & 0.2000 & +0.275 & 0.1451 & 0.2002 & +0.275 & 0.1437 & 0.1918 & +0.251 & 0.1427 & 0.1909 & +0.253 \\
PSCC-Net & 0.1110 & 0.1500 & +0.260 & 0.1115 & 0.1503 & +0.258 & 0.1164 & 0.1484 & +0.215 & 0.1146 & 0.1449 & +0.209 \\
IML-ViT & 0.1300 & 0.1790 & +0.274 & 0.1305 & 0.1795 & +0.273 & 0.1272 & 0.1746 & +0.272 & 0.1253 & 0.1719 & +0.271 \\
TruFor & 0.1050 & 0.1560 & +0.327 & 0.1055 & 0.1567 & +0.327 & 0.1032 & 0.1507 & +0.315 & 0.1017 & 0.1477 & +0.312 \\
SIDA-7B & 0.0260 & 0.0450 & +0.422 & 0.0261 & 0.0453 & +0.424 & 0.0241 & 0.0441 & +0.453 & 0.0227 & 0.0443 & +0.487 \\
SparseViT & 0.1070 & 0.1450 & +0.262 & 0.1072 & 0.1454 & +0.263 & 0.1035 & 0.1437 & +0.279 & 0.1010 & 0.1424 & +0.291 \\
FMAE & 0.1900 & 0.2320 & +0.181 & 0.1901 & 0.2335 & +0.186 & 0.1889 & 0.2347 & +0.195 & 0.1833 & 0.2310 & +0.206 \\
ForgeryTalker & 0.1340 & 0.1760 & +0.239 & 0.1343 & 0.1760 & +0.237 & 0.1344 & 0.1757 & +0.235 & 0.1358 & 0.1745 & +0.222 \\
\bottomrule\end{tabular*}
\end{table*}

\section{Mask-Area Matching and Localization F1}
\label{app:area_matched}
We use the mask-area fractions $A_i^z$ and paired area similarity $s_{\rm area}$ defined in Section~\ref{sec:magnitude_controls}, Eq.~\ref{eq:area-similarity}.
We retain common official-test pairs with $s_{\rm area}\geq0.9$ and evaluate all nine localizers on identical pair IDs. The prediction threshold is 0.5. This area filter is applied independently of the pixel-change filter. It retains 14,511 of 34,543 official-test pairs (42.0\%).

\begin{table}[htbp]\centering\footnotesize
\caption{Official-test size composition and paired area balance before and after mask-area matching. Size shares use each selected subset. MAD is the mean absolute paired area difference.}
\label{tab:area-match-balance}
\setlength{\tabcolsep}{4pt}
\begin{tabular}{@{}lcc@{}}\toprule
Metric & All pairs & $s_{\rm area}\geq0.9$ \\\midrule
Retained & 100.0\% & 42.0\% \\
Small & 4.3\% & 3.9\% \\
Medium & 30.2\% & 26.4\% \\
Large & 65.4\% & 69.7\% \\
Mean area P / A & 0.3186 / 0.3407 & 0.3661 / 0.3687 \\
Area MAD & 0.0579 & 0.0144 \\
\bottomrule\end{tabular}
\end{table}
The shared size label uses $\bar A_i=(A_i^\pl+A_i^\an)/2$. Small regions have $\bar A_i<0.05$, medium regions have $0.05\leq\bar A_i<0.20$, and large regions have $\bar A_i\geq0.20$. Matching reduces mean absolute paired area difference from 0.0579 to 0.0144. The large-region share increases from 65.4\% to 69.7\%. We therefore report the matched pairs separately within each size bin. Gaps remain positive in 26/27 small, 26/27 medium, and 23/27 large comparisons, totaling 75/81 across nine localizers.

\begin{table*}[tp]\centering\footnotesize
\caption{Nine-model localization on COCO after mask-area matching ($s_{\rm area}\geq0.9$), separated by region size.}
\label{tab:area-match-coco}
\setlength{\tabcolsep}{2.5pt}\renewcommand{\arraystretch}{1.04}
\begin{tabular*}{\textwidth}{@{\extracolsep{\fill}}l*{9}{r}@{}}\toprule
 & \multicolumn{3}{c}{Small} & \multicolumn{3}{c}{Medium} & \multicolumn{3}{c}{Large} \\
\cmidrule(lr){2-4}\cmidrule(lr){5-7}\cmidrule(lr){8-10}
Method & $J_\pl$ & $J_\an$ & $\Delta_{\rm rel}$ & $J_\pl$ & $J_\an$ & $\Delta_{\rm rel}$ & $J_\pl$ & $J_\an$ & $\Delta_{\rm rel}$ \\\midrule
CAT-Net & 0.2604 & 0.2744 & +0.051 & 0.0681 & 0.0506 & -0.346 & 0.0458 & 0.0239 & -0.922 \\
ObjectFormer & 0.1151 & 0.2341 & +0.508 & 0.1838 & 0.2125 & +0.135 & 0.2081 & 0.2074 & -0.003 \\
PSCC-Net & 0.1752 & 0.2620 & +0.331 & 0.1446 & 0.1594 & +0.093 & 0.1301 & 0.1286 & -0.012 \\
IML-ViT & 0.2452 & 0.4567 & +0.463 & 0.1604 & 0.2038 & +0.213 & 0.1007 & 0.1218 & +0.173 \\
TruFor & 0.1451 & 0.3424 & +0.576 & 0.1048 & 0.1383 & +0.242 & 0.0387 & 0.0527 & +0.266 \\
SIDA-7B & 0.0085 & 0.0303 & +0.720 & 0.0288 & 0.0355 & +0.190 & 0.0223 & 0.0574 & +0.611 \\
SparseViT & 0.1946 & 0.2633 & +0.261 & 0.0837 & 0.1427 & +0.413 & 0.0481 & 0.0657 & +0.267 \\
FMAE & 0.2400 & 0.3578 & +0.329 & 0.1493 & 0.1720 & +0.132 & 0.1967 & 0.2034 & +0.033 \\
ForgeryTalker & 0.0281 & 0.0541 & +0.481 & 0.1089 & 0.1221 & +0.108 & 0.1707 & 0.2154 & +0.207 \\
\bottomrule\end{tabular*}
\end{table*}

\begin{table*}[tp]\centering\footnotesize
\caption{Nine-model localization on ImageNet after mask-area matching ($s_{\rm area}\geq0.9$), separated by region size.}
\label{tab:area-match-imagenet}
\setlength{\tabcolsep}{2.5pt}\renewcommand{\arraystretch}{1.04}
\begin{tabular*}{\textwidth}{@{\extracolsep{\fill}}l*{9}{r}@{}}\toprule
 & \multicolumn{3}{c}{Small} & \multicolumn{3}{c}{Medium} & \multicolumn{3}{c}{Large} \\
\cmidrule(lr){2-4}\cmidrule(lr){5-7}\cmidrule(lr){8-10}
Method & $J_\pl$ & $J_\an$ & $\Delta_{\rm rel}$ & $J_\pl$ & $J_\an$ & $\Delta_{\rm rel}$ & $J_\pl$ & $J_\an$ & $\Delta_{\rm rel}$ \\\midrule
CAT-Net & 0.1618 & 0.2002 & +0.192 & 0.0814 & 0.1090 & +0.254 & 0.0465 & 0.0473 & +0.017 \\
ObjectFormer & 0.1380 & 0.1702 & +0.189 & 0.1967 & 0.2118 & +0.071 & 0.1675 & 0.1817 & +0.078 \\
PSCC-Net & 0.1811 & 0.2034 & +0.110 & 0.1905 & 0.1972 & +0.034 & 0.1485 & 0.1484 & -0.001 \\
IML-ViT & 0.3171 & 0.3438 & +0.078 & 0.2231 & 0.2468 & +0.096 & 0.1245 & 0.1457 & +0.146 \\
TruFor & 0.2960 & 0.2747 & -0.078 & 0.1368 & 0.1731 & +0.210 & 0.0570 & 0.0652 & +0.125 \\
SIDA-7B & 0.0174 & 0.0211 & +0.177 & 0.0252 & 0.0396 & +0.364 & 0.0224 & 0.0310 & +0.276 \\
SparseViT & 0.2257 & 0.2335 & +0.033 & 0.1460 & 0.1691 & +0.137 & 0.0675 & 0.0741 & +0.089 \\
FMAE & 0.2301 & 0.2610 & +0.118 & 0.2984 & 0.3240 & +0.079 & 0.2617 & 0.2783 & +0.059 \\
ForgeryTalker & 0.0456 & 0.0481 & +0.051 & 0.1146 & 0.1460 & +0.215 & 0.2358 & 0.2786 & +0.153 \\
\bottomrule\end{tabular*}
\end{table*}

\begin{table*}[tp]\centering\footnotesize
\caption{Nine-model localization on Places after mask-area matching ($s_{\rm area}\geq0.9$), separated by region size.}
\label{tab:area-match-places}
\setlength{\tabcolsep}{2.5pt}\renewcommand{\arraystretch}{1.04}
\begin{tabular*}{\textwidth}{@{\extracolsep{\fill}}l*{9}{r}@{}}\toprule
 & \multicolumn{3}{c}{Small} & \multicolumn{3}{c}{Medium} & \multicolumn{3}{c}{Large} \\
\cmidrule(lr){2-4}\cmidrule(lr){5-7}\cmidrule(lr){8-10}
Method & $J_\pl$ & $J_\an$ & $\Delta_{\rm rel}$ & $J_\pl$ & $J_\an$ & $\Delta_{\rm rel}$ & $J_\pl$ & $J_\an$ & $\Delta_{\rm rel}$ \\\midrule
CAT-Net & 0.1560 & 0.2202 & +0.292 & 0.0651 & 0.0940 & +0.308 & 0.0410 & 0.0514 & +0.202 \\
ObjectFormer & 0.1176 & 0.1315 & +0.105 & 0.1608 & 0.2057 & +0.218 & 0.1393 & 0.1809 & +0.230 \\
PSCC-Net & 0.1442 & 0.1778 & +0.189 & 0.1285 & 0.1635 & +0.214 & 0.1041 & 0.1264 & +0.176 \\
IML-ViT & 0.2783 & 0.2806 & +0.008 & 0.1651 & 0.2272 & +0.273 & 0.0880 & 0.1154 & +0.238 \\
TruFor & 0.2254 & 0.2999 & +0.248 & 0.1475 & 0.2128 & +0.307 & 0.0540 & 0.0901 & +0.401 \\
SIDA-7B & 0.0270 & 0.0447 & +0.395 & 0.0312 & 0.0580 & +0.462 & 0.0218 & 0.0362 & +0.397 \\
SparseViT & 0.1926 & 0.2469 & +0.220 & 0.1419 & 0.1898 & +0.252 & 0.0642 & 0.0869 & +0.261 \\
FMAE & 0.2300 & 0.2649 & +0.132 & 0.2255 & 0.2824 & +0.202 & 0.1601 & 0.1934 & +0.172 \\
ForgeryTalker & 0.0324 & 0.0355 & +0.086 & 0.0875 & 0.1087 & +0.195 & 0.1849 & 0.2315 & +0.201 \\
\bottomrule\end{tabular*}
\end{table*}

\subsection{Localization F1 on the full test cohort}
We also evaluate pixel-level $F_1(B,M)=2|B\cap M|/(|B|+|M|)$ on all official-test pairs. Its relative gap is $(\bar F_{1,\an}-\bar F_{1,\pl})/\bar F_{1,\an}$. Table~\ref{tab:f1} has positive gaps in 26/27 model--dataset comparisons, with CAT-Net on COCO the exception.

\begin{table*}[tp]\centering\footnotesize
\caption{\rev{Localization F1 and its relative gap.}}\label{tab:f1}
\setlength{\tabcolsep}{2.5pt}\renewcommand{\arraystretch}{1.04}
\begin{tabular*}{\textwidth}{@{\extracolsep{\fill}}l*{9}{r}@{}}\toprule
 & \multicolumn{3}{c}{COCO} & \multicolumn{3}{c}{ImageNet} & \multicolumn{3}{c}{Places} \\
\cmidrule(lr){2-4}\cmidrule(lr){5-7}\cmidrule(lr){8-10}
Method & $F_{1,\pl}$ & $F_{1,\an}$ & $\Delta^{F_1}_{\rm rel}$ & $F_{1,\pl}$ & $F_{1,\an}$ & $\Delta^{F_1}_{\rm rel}$ & $F_{1,\pl}$ & $F_{1,\an}$ & $\Delta^{F_1}_{\rm rel}$ \\\midrule
CAT-Net & 0.072 & 0.070 & -0.027 & 0.083 & 0.092 & 0.102 & 0.073 & 0.112 & 0.351 \\
ObjectFormer & 0.246 & 0.291 & 0.156 & 0.232 & 0.262 & 0.114 & 0.194 & 0.265 & 0.267 \\
PSCC-Net & 0.193 & 0.207 & 0.069 & 0.223 & 0.230 & 0.032 & 0.158 & 0.209 & 0.247 \\
IML-ViT & 0.192 & 0.231 & 0.168 & 0.219 & 0.246 & 0.111 & 0.178 & 0.240 & 0.260 \\
TruFor & 0.087 & 0.119 & 0.267 & 0.100 & 0.121 & 0.179 & 0.127 & 0.190 & 0.331 \\
SIDA-7B & 0.036 & 0.059 & 0.382 & 0.027 & 0.042 & 0.360 & 0.033 & 0.058 & 0.436 \\
SparseViT & 0.095 & 0.133 & 0.283 & 0.116 & 0.133 & 0.132 & 0.133 & 0.181 & 0.264 \\
FMAE & 0.234 & 0.265 & 0.115 & 0.359 & 0.382 & 0.061 & 0.248 & 0.306 & 0.190 \\
ForgeryTalker & 0.202 & 0.253 & 0.202 & 0.282 & 0.341 & 0.174 & 0.208 & 0.266 & 0.217 \\
\bottomrule\end{tabular*}
\end{table*}

\section{Localization with Pixel-Difference Masks}
\label{app:diff_mask}
We test whether the plausible-edit deficit depends on the instruction-conditioned SIGMA reference masks. We construct an alternative binary mask for each edited output by thresholding its per-pixel mean absolute RGB difference from the source at 0.5, with RGB values normalized to $[0,1]$. This construction uses only the source and edited images. We reuse the existing predictions of nine localizers, keep the prediction threshold at 0.5, and rescore them against these alternative reference masks on the official-test pairs. Within each dataset, we average the two condition-specific IoUs equally over the four editors and then compute the relative gap using main-paper Eq.~\ref{eq:gap}.

\paragraph{Results and interpretation.}
Table~\ref{tab:diff-mask} reports the condition-specific IoUs and relative gaps on COCO, ImageNet, and Places. Plausible-edit IoU is lower in 26/27 model--dataset comparisons: eight of nine methods on COCO and all nine on ImageNet and Places. CAT-Net on COCO is the sole exception, with low plausible/anomalous IoUs of 0.022/0.018 and a relative gap of $-0.228$. Gap magnitudes vary across methods, from $+0.013$ for PSCC-Net to $+0.511$ for SIDA-7B on COCO. The recurring plausible-edit deficit under these instruction-independent masks shows that the trend also appears with a reference-mask construction that does not use SIGMA or the edit instruction.

\begin{table*}[tp]\centering\footnotesize
\caption{Localization with source--output pixel-difference masks for nine localizers. Reference masks are obtained by thresholding normalized per-pixel mean absolute RGB difference at 0.5. The prediction threshold is also 0.5. Condition-specific IoUs are averaged equally over editors within each dataset, and relative gaps are computed before rounding.}
\label{tab:diff-mask}
\setlength{\tabcolsep}{2.5pt}\renewcommand{\arraystretch}{1.04}
\begin{tabular*}{\textwidth}{@{\extracolsep{\fill}}l*{9}{r}@{}}\toprule
 & \multicolumn{3}{c}{COCO} & \multicolumn{3}{c}{ImageNet} & \multicolumn{3}{c}{Places} \\
\cmidrule(lr){2-4}\cmidrule(lr){5-7}\cmidrule(lr){8-10}
Method & $J_\pl$ & $J_\an$ & $\Delta_{\rm rel}$ & $J_\pl$ & $J_\an$ & $\Delta_{\rm rel}$ & $J_\pl$ & $J_\an$ & $\Delta_{\rm rel}$ \\\midrule
CAT-Net & 0.022 & 0.018 & -0.228 & 0.026 & 0.028 & 0.090 & 0.020 & 0.031 & 0.357 \\
ObjectFormer & 0.069 & 0.076 & 0.096 & 0.066 & 0.074 & 0.110 & 0.048 & 0.065 & 0.272 \\
PSCC-Net & 0.048 & 0.049 & 0.013 & 0.054 & 0.059 & 0.085 & 0.033 & 0.047 & 0.300 \\
IML-ViT & 0.058 & 0.065 & 0.106 & 0.060 & 0.073 & 0.177 & 0.047 & 0.068 & 0.309 \\
TruFor & 0.026 & 0.028 & 0.049 & 0.028 & 0.033 & 0.158 & 0.030 & 0.047 & 0.372 \\
SIDA-7B & 0.008 & 0.016 & 0.511 & 0.008 & 0.011 & 0.281 & 0.008 & 0.013 & 0.371 \\
SparseViT & 0.027 & 0.035 & 0.232 & 0.034 & 0.038 & 0.115 & 0.032 & 0.046 & 0.306 \\
FMAE & 0.066 & 0.074 & 0.097 & 0.100 & 0.109 & 0.085 & 0.066 & 0.083 & 0.204 \\
ForgeryTalker & 0.041 & 0.051 & 0.192 & 0.062 & 0.074 & 0.160 & 0.038 & 0.048 & 0.210 \\
\bottomrule\end{tabular*}
\end{table*}
\section{Human Visual Assessment}
\label{app:human}

\paragraph{Localization by verified realization.}
The recorded human judgment partitions the same 500 pairs into 372 realized and 128 unrealized violations. Each anomalous output keeps its original plausible counterpart. We rescore the existing predictions within each group. Main-paper Table~\ref{tab:realization} reports the two groups using the paper's evaluation protocol. All nine public localizers retain positive gaps on realized violations. Only SIDA-7B, ForgeryTalker, and TruFor do so on unrealized violations.

\begin{table}[htbp]\centering\footnotesize
\caption{Human visual assessment of 500 English pairs. In two-line cells, plausible results appear above anomalous results.}
\label{tab:human}
\setlength{\tabcolsep}{2.5pt}
\begin{tabular*}{\columnwidth}{@{\extracolsep{\fill}}lcccc@{}}\toprule
Metric & COCO & ImageNet & Places & All \\\midrule
Pairs & 167 & 167 & 166 & 500 \\
Mean mask score & \shortstack{4.898\\4.856} & \shortstack{4.623\\4.587} & \shortstack{4.235\\4.283} & \shortstack{4.586\\4.576} \\
Rated $\geq4$ & \shortstack{98.2\%\\96.4\%} & \shortstack{94.0\%\\93.4\%} & \shortstack{79.5\%\\83.1\%} & \shortstack{90.6\%\\91.0\%} \\
Violation realized & \shortstack{65.3\%\\(109/167)} & \shortstack{82.6\%\\(138/167)} & \shortstack{75.3\%\\(125/166)} & \shortstack{74.4\%\\(372/500)} \\
Visual near-copy & \shortstack{0.0\%\\(0/167)} & \shortstack{0.0\%\\(0/167)} & \shortstack{0.0\%\\(0/166)} & \shortstack{0.0\%\\(0/500)} \\
\bottomrule\end{tabular*}
\end{table}

\rev{As shown in Table~\ref{tab:human}, plausible and anomalous masks receive mean scores of 4.586 and 4.576, with 90.6\% and 91.0\% rated at least 4. Ratings are equal within 403 pairs (80.6\%), higher for the plausible mask in 52 (10.4\%), and higher for the anomalous mask in 45 (9.0\%). These ratings indicate close correspondence to visible edits and similar mask quality across conditions. Experts confirm the assigned violation in 372/500 anomalous images (74.4\%) and find no visually near-copy plausible edit. This visual near-copy judgment is separate from the pixel threshold above.}

\begin{figure*}[tp]\centering
\includegraphics[width=\textwidth]{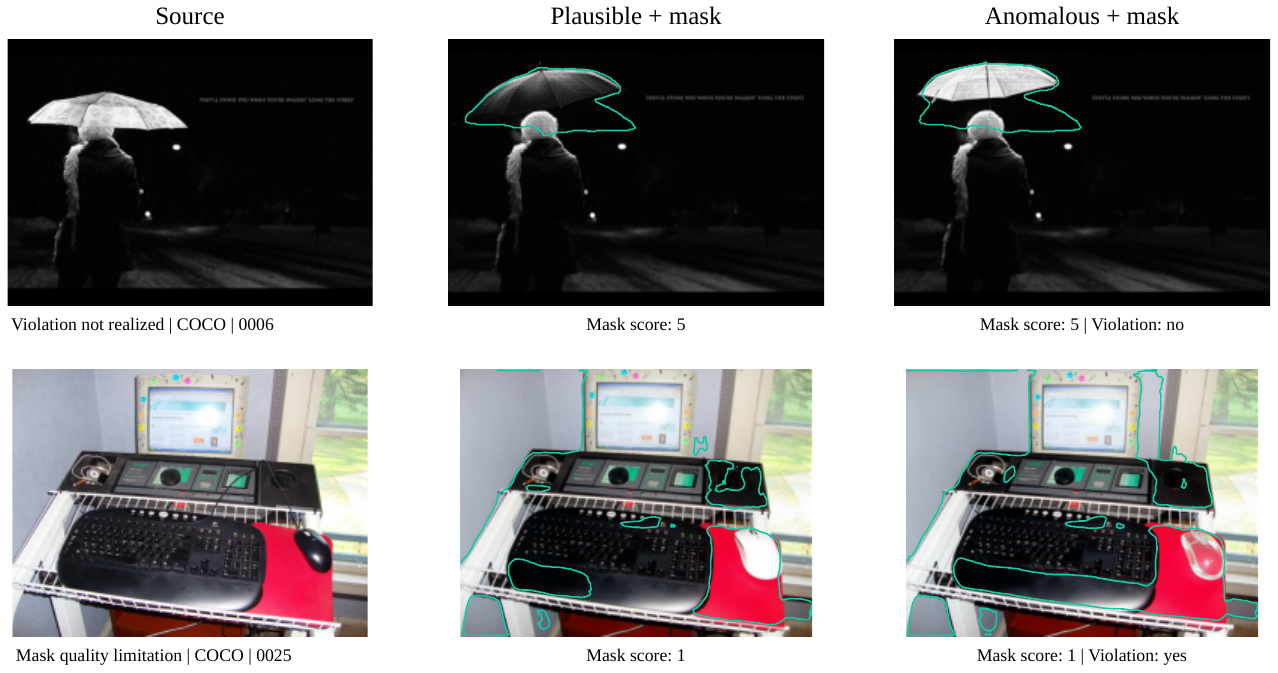}
\caption{\rev{Audited examples of an unrealized violation (top) and low-quality masks (bottom), with their recorded ratings and realization judgments.}}\label{fig:audit-failures}
\end{figure*}

This appendix supplements Section~\ref{sec:human} with visual examples of the human assessment. Figure~\ref{fig:audit-success} covers all six violation types, while Figure~\ref{fig:audit-failures} shows an unrealized violation and low-quality masks. The source images, edited outputs, mask contours, and recorded judgments allow direct comparison with the visible changes.

\paragraph{Joint mask quality within pairs.}
Both masks receive scores of at least 4 in 439 of the 500 audited pairs (87.8\%). This complements the condition-specific scores in Table~\ref{tab:human} by reporting how often both members of a pair have highly rated masks.

\begin{figure*}[tp]\centering
\input{figures/human_audit_examples_compact}
\caption{\rev{Human-audited examples covering the six violation types. Each labeled group shows the source, plausible output, and anomalous output from left to right, with mask contours on the outputs. Both outputs in each pair received a mask score of 5, and all six assigned violations were judged as realized.}}\label{fig:audit-success}
\end{figure*}

\section{Stylized-Image Extension}
\label{app:stylized}
Stylized images have different textures and visual conventions from photographs. We evaluate paired cartoon edits to test whether the plausible-edit deficit also appears in this domain. Table~\ref{tab:stylized} compares plausible and anomalous localization for all nine methods on a cartoon-image collection, using the same paired evaluation protocol.

We randomly sampled 1,000 images from \href{https://huggingface.co/datasets/Norod78/cartoon-blip-captions}{\texttt{norod78/cartoon-blip-captions}} on Hugging Face and generated paired replacements with Qwen-Image-Edit using the same prompt-generation, mask-generation, and filtering pipeline. The evaluation contains 1,341 pairs that pass the pair-mask IoU requirement of at least 0.5.

\paragraph{Localization on stylized edits.}
All nine localizers have positive relative gaps, ranging from 0.081 to 0.572 (Table~\ref{tab:stylized}). FMAE has the highest plausible-edit IoU, 0.279, and still obtains a higher anomalous-edit IoU of 0.306, giving a relative gap of 0.087. SparseViT obtains IoUs of 0.150/0.190 and a gap of 0.214. The plausible-edit deficit therefore also occurs in this cartoon collection under Qwen editing, across methods with different levels of absolute localization accuracy.

\begin{table}[htbp]\centering\small

\caption{\rev{Localization on 1,341 stylized-image pairs generated with Qwen-Image-Edit.}}\label{tab:stylized}

\setlength{\tabcolsep}{4pt}\renewcommand{\arraystretch}{1.12}

\begin{lrbox}{\revisiontablebox}

\begin{tabular}{@{}lccc@{}}\toprule

Method & $J_\pl$ & $J_\an$ & $\Delta_{\rm rel}$ \\\midrule

CAT-Net & 0.094 & 0.112 & 0.164 \\
ObjectFormer & 0.163 & 0.194 & 0.161 \\
PSCC-Net & 0.121 & 0.162 & 0.251 \\
IML-ViT & 0.107 & 0.152 & 0.299 \\
TruFor & 0.160 & 0.222 & 0.281 \\
SIDA-7B & 0.006 & 0.015 & 0.572 \\
SparseViT & 0.150 & 0.190 & 0.214 \\
FMAE & 0.279 & 0.306 & 0.087 \\
ForgeryTalker & 0.198 & 0.215 & 0.081 \\

\bottomrule\end{tabular}\end{lrbox}

\ifdim\wd\revisiontablebox>\linewidth\resizebox{\linewidth}{!}{\usebox{\revisiontablebox}}\else\usebox{\revisiontablebox}\fi

\end{table}

\section{Non-Edited Anomalous Images}
\label{app:unedited}

\begin{figure*}[!tp]\centering
\includegraphics[width=\textwidth]{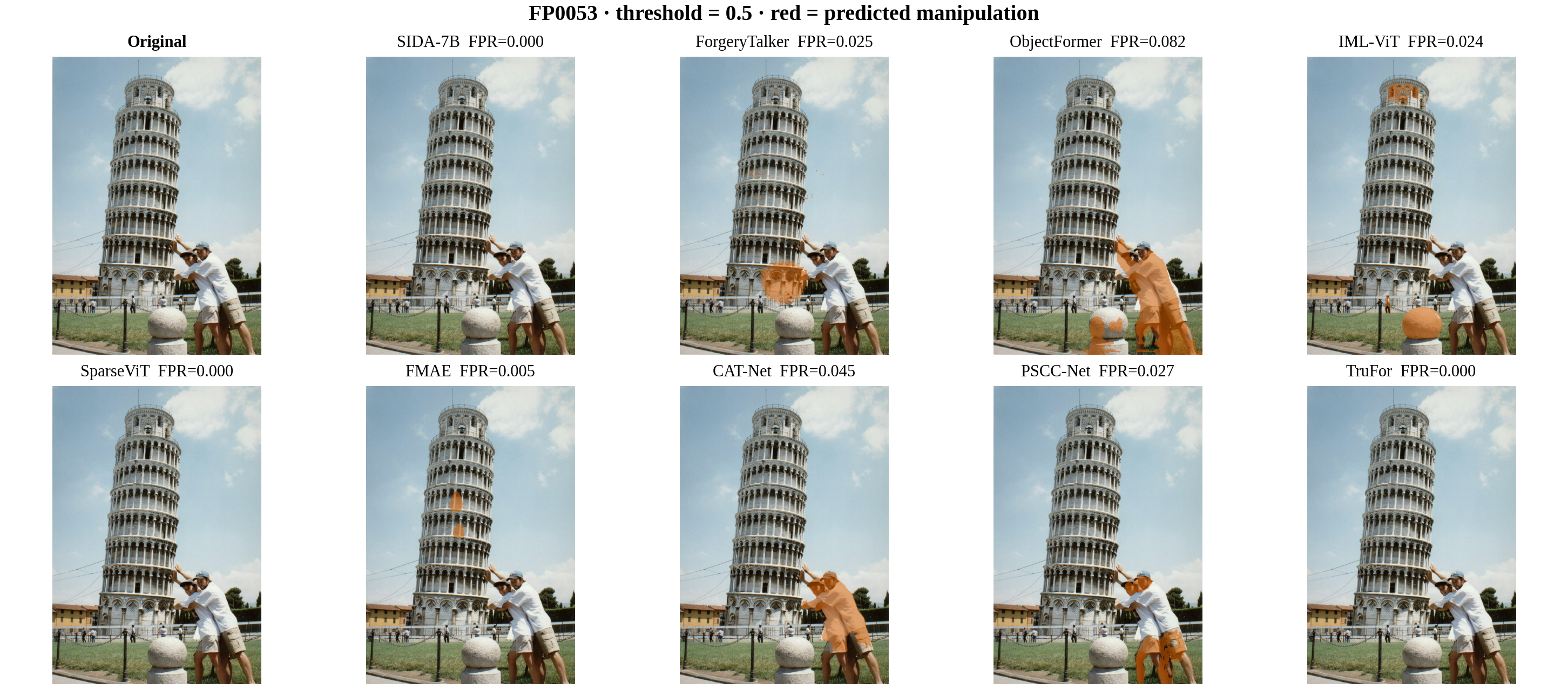}
\caption{\rev{Non-edited anomalous photograph FP0053 and predictions from all nine public localizers at threshold 0.5.}}\label{fig:unedited-1}
\end{figure*}

\begin{figure*}[!tp]\centering
\includegraphics[width=\textwidth]{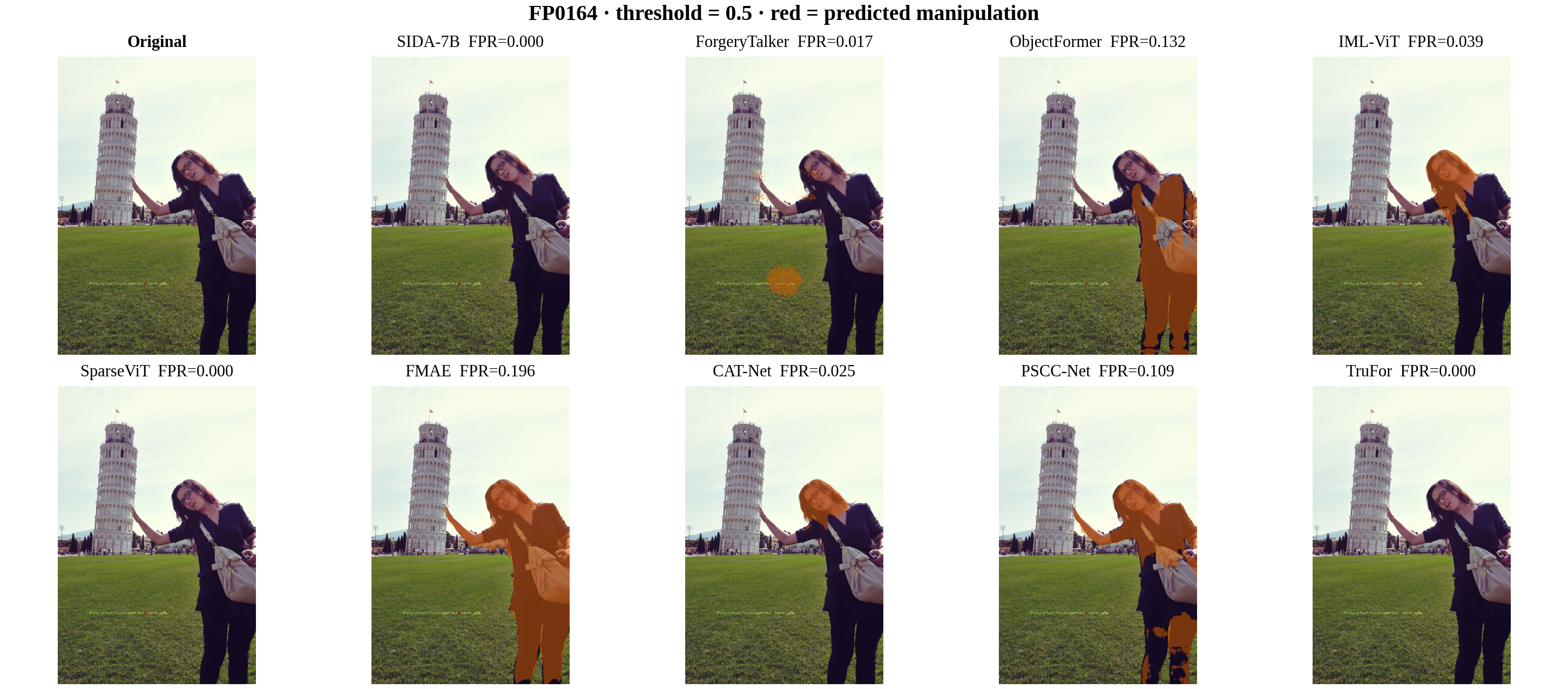}
\caption{\rev{Non-edited anomalous photograph FP0164 and predictions from all nine public localizers at threshold 0.5.}}\label{fig:unedited-2}
\end{figure*}

\begin{figure*}[!tp]\centering
\includegraphics[width=\textwidth]{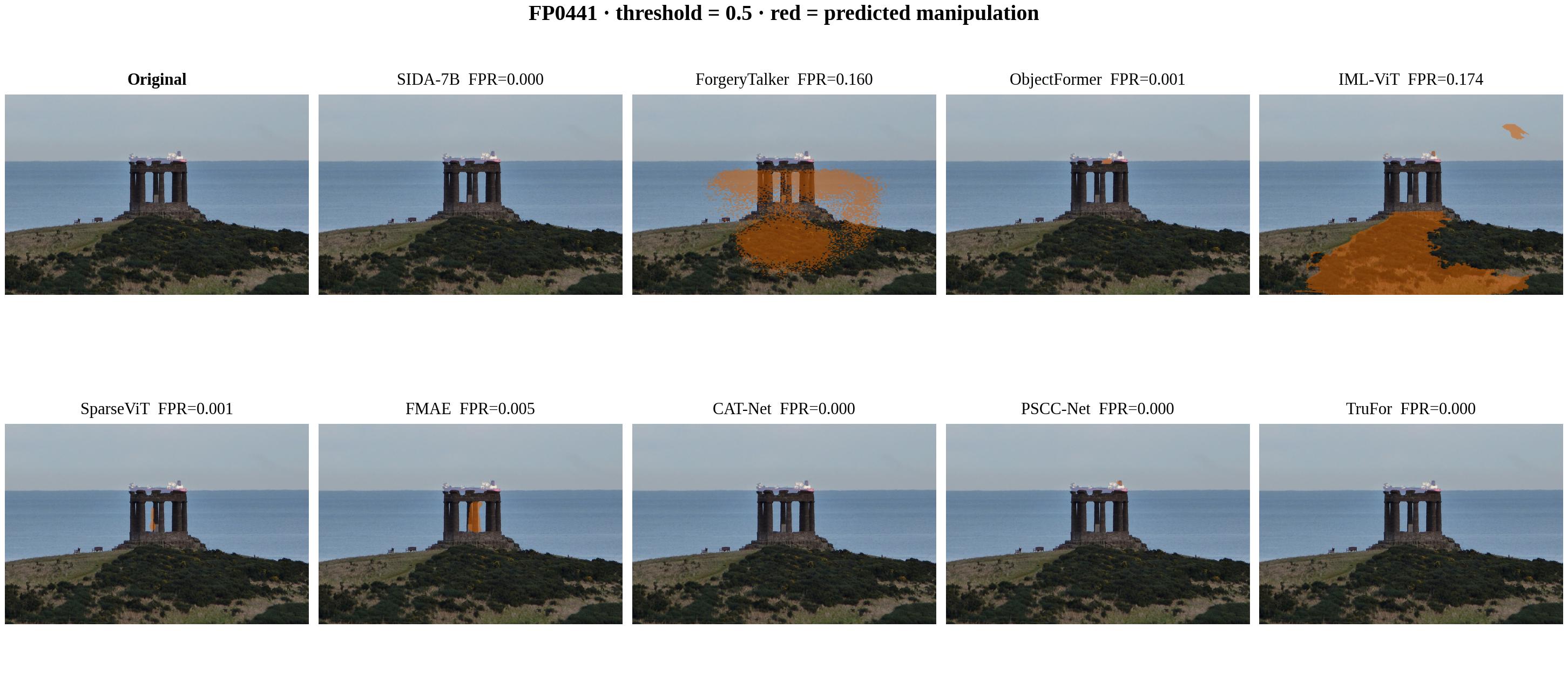}
\caption{\rev{Non-edited anomalous photograph FP0441 and predictions from all nine public localizers at threshold 0.5.}}\label{fig:unedited-3}
\end{figure*}

The paired IoU evaluation measures localization on edited images. This appendix examines how localizers respond to anomalous-looking photographs with no edited region. Table~\ref{tab:unedited-fpr} reports the extent and frequency of false-positive localization at fixed prediction thresholds, while Figures~\ref{fig:unedited-1}--\ref{fig:unedited-3} show where the models respond on reviewed examples.

We collected photographs from public photography communities, focusing on unusual apparent scales and interactions caused by perspective, occlusion, and viewpoint. Manual image and source-page review excluded generated or composited images and photographs with weak anomalous relationships. Exact and perceptual-hash deduplication yielded 500 non-edited anomalous images. With an empty manipulation reference, every predicted positive pixel is a false positive. For score map $S_i$ with $H_iW_i$ output-grid pixels, we measure
\begin{align}
\mathrm{FPR}_{\rm pixel}(\tau)&=\frac{1}{N}\sum_i\frac{\sum_p\mathbb{1}[S_i(p)>\tau]}{H_iW_i},\\
\mathrm{FPR}_{\rm image}(\tau)&=\frac{1}{N}\sum_i\mathbb{1}[\max_p S_i(p)>\tau].
\end{align}

\paragraph{Extent of false-positive localization.}
At threshold 0.5, mean false-positive pixel area ranges from 0.37\% for SIDA-7B to 11.99\% for ObjectFormer (Table~\ref{tab:unedited-fpr}). FMAE and PSCC-Net also assign more than 11\% of pixels to manipulation on average. These results quantify substantial false-positive regions on part of this non-edited photograph collection. Figures~\ref{fig:unedited-1}--\ref{fig:unedited-3} show the corresponding spatial responses.

\paragraph{Frequency of image-level false alarms.}
The proportion of images with at least one positive pixel ranges from 1.8\% to 100.0\% at threshold 0.5 (Table~\ref{tab:unedited-fpr}). SparseViT flags 63.2\% of images despite a mean false-positive area of 1.49\%. Thus, image-level frequency and pixel-level extent capture different aspects of the errors: localized responses can occur on many images even when their average area is small.

\paragraph{Sensitivity to the prediction threshold.}
Raising the threshold reduces false-positive area, but every localizer retains a nonzero mean false-positive area at 0.9 (Table~\ref{tab:unedited-fpr}). False alarms on this collection therefore persist at stricter operating thresholds.

\begin{table}[!htbp]\centering\footnotesize
\caption{False positives on 500 non-edited anomalous images. Pixel rates average per-image output-grid fractions.}
\label{tab:unedited-fpr}
\setlength{\tabcolsep}{0.8pt}\renewcommand{\arraystretch}{1.04}
\begin{tabular*}{\columnwidth}{@{\extracolsep{\fill}}l*{6}{r}@{}}\toprule
Method & Pixel .5 & Pixel .6 & Pixel .7 & Pixel .8 & Pixel .9 & Image .5 \\\midrule
CAT-Net & 1.85\% & 1.57\% & 1.30\% & 0.97\% & 0.63\% & 61.4\% \\
ObjectFormer & 11.99\% & 11.60\% & 11.19\% & 10.70\% & 10.00\% & 94.0\% \\
PSCC-Net & 11.17\% & 9.17\% & 7.23\% & 5.26\% & 3.11\% & 97.0\% \\
IML-ViT & 3.96\% & 3.49\% & 3.03\% & 2.52\% & 1.87\% & 100.0\% \\
TruFor & 1.77\% & 1.55\% & 1.34\% & 1.14\% & 0.89\% & 46.6\% \\
SIDA-7B & 0.37\% & 0.33\% & 0.25\% & 0.15\% & 0.11\% & 1.8\% \\
SparseViT & 1.49\% & 1.10\% & 0.77\% & 0.52\% & 0.33\% & 63.2\% \\
FMAE & 11.52\% & 10.36\% & 8.75\% & 6.60\% & 3.17\% & 99.6\% \\
ForgeryTalker & 6.69\% & 4.62\% & 2.98\% & 1.70\% & 0.70\% & 100.0\% \\
\bottomrule\end{tabular*}
\end{table}

\section{Commercial Closed-Source Localization}
\label{app:textin}
We extend the paired evaluation to TextIn to examine whether a commercial localization service exhibits the same plausible-edit deficit as the public checkpoints. Table~\ref{tab:textin} compares TextIn with all nine public localizers on the same 500 audited English pairs, so both semantic conditions and the evaluation cohort are shared across methods.

We evaluated TextIn, the commercial image-manipulation detection service from INTSIG. Its returned PS-tamper quadrilaterals and confidence scores are rasterized into localization maps, thresholded at 0.5, and scored against each output's SIGMA mask.

\paragraph{Paired localization results.}
TextIn obtains plausible/anomalous mean IoUs of 0.072/0.122 and a relative gap of 0.412 (Table~\ref{tab:textin}). All ten methods have positive gaps on this matched subset. The lower plausible-edit IoU therefore also appears in this commercial service, alongside the public localizers evaluated on the same images.

\begin{table}[htbp]\centering\footnotesize

\caption{\rev{Public checkpoints and TextIn on the same 500-pair human-audit subset.}}\label{tab:textin}

\setlength{\tabcolsep}{2pt}\renewcommand{\arraystretch}{1.12}

\begin{lrbox}{\revisiontablebox}

\begin{tabular}{@{}lcccc@{}}\toprule

Method & Access & $J_\pl$ & $J_\an$ & $\Delta_{\rm rel}$ \\\midrule

CAT-Net & Public checkpoint & 0.053 & 0.064 & 0.171 \\

ObjectFormer & Public checkpoint & 0.152 & 0.202 & 0.248 \\

PSCC-Net & Public checkpoint & 0.139 & 0.156 & 0.110 \\

IML-ViT & Public checkpoint & 0.147 & 0.181 & 0.190 \\

TruFor & Public checkpoint & 0.085 & 0.135 & 0.372 \\

SIDA-7B & Public checkpoint & 0.006 & 0.027 & 0.768 \\

SparseViT & Public checkpoint & 0.090 & 0.113 & 0.200 \\

FMAE & Public checkpoint & 0.202 & 0.224 & 0.100 \\

ForgeryTalker & Public checkpoint & 0.129 & 0.204 & 0.369 \\

TextIn & Closed-source API & 0.072 & 0.122 & 0.412 \\

\bottomrule\end{tabular}\end{lrbox}

\ifdim\wd\revisiontablebox>\linewidth\resizebox{\linewidth}{!}{\usebox{\revisiontablebox}}\else\usebox{\revisiontablebox}\fi

\end{table}

\end{document}